\documentclass{article}

\usepackage{microtype}
\usepackage{graphicx}
\usepackage{subfigure}
\usepackage{booktabs} 
\usepackage{setspace}

\usepackage[accepted,nohyperref]{icml2024}

\usepackage{amsmath}
\usepackage{amssymb}
\usepackage{mathtools}
\usepackage{amsthm}
\usepackage{dsfont}  
\usepackage{tikz}

\usepackage[hidelinks]{hyperref}       
\definecolor{mydarkblue}{rgb}{0,0.08,0.45}
\hypersetup{colorlinks,linkcolor={mydarkblue},citecolor={mydarkblue},urlcolor={red}}

\usepackage[capitalize,noabbrev]{cleveref}

\usepackage{natbib}
\setcitestyle{authoryear,round,citesep={,},aysep={},yysep={-}}
\newcommand{\citenoh}[1]{{\protect\NoHyper\cite{#1}\protect\endNoHyper}}
\newcommand{\citetnoh}[1]{{\protect\NoHyper\citet{#1}\protect\endNoHyper}}
\newcommand{\nocitenoh}[1]{{\protect\NoHyper\nocite{#1}\protect\endNoHyper}}
\definecolor{darkblue}{rgb}{0, 0, 0.55}

\theoremstyle{plain}
\newtheorem{theorem}{Theorem}[section]
\newtheorem{proposition}[theorem]{Proposition}
\newtheorem{lemma}[theorem]{Lemma}
\newtheorem{corollary}[theorem]{Corollary}
\theoremstyle{definition}

\theoremstyle{remark}
\newtheorem{remark}[theorem]{Remark}

\DeclarePairedDelimiter{\ceil}{\lceil}{\rceil}
\usepackage{tabularx}  
\usepackage{ragged2e}  
\newcolumntype{Y}{>{\RaggedRight\arraybackslash}X} 
\usepackage{perpage}

\usepackage[textsize=tiny]{todonotes}

\icmltitlerunning{Target Networks and Over-parameterization Stabilize Off-policy Bootstrapping}

\begin{document}

\twocolumn[
\icmltitle{
Target Networks and Over-parameterization Stabilize Off-policy Bootstrapping with Function Approximation
}




\begin{icmlauthorlist}
\icmlauthor{Fengdi Che}{yyy}
\icmlauthor{Chenjun Xiao}{xxx}
\icmlauthor{Jincheng Mei}{deepmind}
\icmlauthor{Bo Dai}{deepmind,zzz}
\icmlauthor{Ramki Gummadi}{deepmind}
\icmlauthor{Oscar A Ramirez}{figure,notice}
\icmlauthor{Christopher K Harris}{notice,uber}
\icmlauthor{A. Rupam Mahmood}{yyy,amii}
\icmlauthor{Dale Schuurmans}{yyy,deepmind,amii}
\end{icmlauthorlist}

\icmlaffiliation{yyy}{Department of Computing Science, University of Alberta}
\icmlaffiliation{xxx}{School of Data Science, The Chinese University of Hong Kong, Shenzhen}
\icmlaffiliation{zzz}{School of Computational Science and Engineering, Georgia Tech}
\icmlaffiliation{deepmind}{Google DeepMind}
\icmlaffiliation{uber}{Uber}
\icmlaffiliation{figure}{Figure}
\icmlaffiliation{amii}{CIFAR AI Chair, Amii}
\icmlaffiliation{notice}{The work was done while the author was at Google.}

\icmlcorrespondingauthor{Chenjun Xiao}{chenjunx@cuhk.edu.cn}

\icmlkeywords{Machine Learning, ICML}

\vskip 0.3in
]

\newcommand{\cS}{\mathcal{S}}
\newcommand{\cD}{\mathcal{D}}
\newcommand{\cA}{\mathcal{A}}
\newcommand{\cX}{\mathcal{X}}
\newcommand{\RR}{\mathbb{R}}
\newcommand{\EE}{\mathbb{E}}



\printAffiliationsAndNotice{Fengdi Che}{fengdi@ualberta.ca} 

\begin{abstract}
We prove that the combination of a target network and over-parameterized linear function approximation establishes a weaker convergence condition for bootstrapped value estimation in certain cases, even with off-policy data.
Our condition is naturally satisfied for expected updates over the entire state-action space or learning with a batch of complete trajectories from episodic Markov decision processes. 
Notably, using only a target network or an over-parameterized model does not provide such a convergence guarantee. 
Additionally, we extend our results to learning with truncated trajectories, showing that convergence is achievable for all tasks with minor modifications, akin to value truncation for the final states in trajectories. 
Our primary result focuses on temporal difference estimation for prediction, providing high-probability value estimation error bounds and empirical analysis on Baird's counterexample and a Four-room task. \footnote{The code is available on the \href{https://github.com/FengdiC/OTTD}{\textcolor{darkblue}{GitHub}}}
Furthermore, we explore the control setting, demonstrating that similar convergence conditions apply to Q-learning.
\end{abstract}

\setlength{\abovedisplayskip}{6.2pt}
\setlength{\abovedisplayshortskip}{2pt}
\setlength{\belowdisplayskip}{6.2pt}
\setlength{\belowdisplayshortskip}{4pt}
\setlength{\jot}{4pt}  
\setlength{\floatsep}{1ex}
\setlength{\textfloatsep}{1ex}

\makeatletter
\renewcommand\paragraph{\@startsection{paragraph}{4}{\z@}{1ex plus
  0ex minus .2ex}{-1em}{\normalsize\bf}}
\makeatother

\section{Introduction}
Off-policy value evaluation with offline data considers the challenge of evaluating the expected discounted cumulative reward of a given policy using a dataset that was not necessarily collected according to the target policy. Such a scenario is common in real-world applications, such as self-driving vehicles and healthcare, where active data collection with an unqualified policy can pose life-threatening risks \citenoh{levine2020offline}. 
The ability to perform off-policy evaluation can also enhance data efficiency, for example, through techniques like experience replay \citenoh{lin1992self}. Typically, estimation from offline data does not enforce constraints on the collection procedure, which allows for the inclusion of diverse sources, such as driving data from multiple drivers or online text from alternative platforms. 
A key challenge, however, is that offline data might only cover part of the state space, which leads to technical difficulties that are not encountered when learning from continual online data.

Temporal difference (TD) estimation \citenoh{sutton2018reinforcement}, where value estimates are formed by bootstrapping from the Bellman equation, has emerged as one of the most widely deployed value estimation techniques. 
Despite its popularity, however, the deadly triad can thwart TD algorithms in offline learning. 
It is well known that the combination of off-policy data, function approximation, and bootstrapping can cause the divergence of such algorithms; 
in the under-parameterized setting, a TD fixed point might not even exist \citenoh{tsitsiklis1996analysis}. 
To address this issue, $L2$-regularization is often introduced to ensure the existence of a fixed point, and several algorithms, such as LSTD \citenoh{yu2009convergence} and TD with a target parameter \citenoh{zhang2021breaking}, have been shown to converge to a regularized fixed point. 
However, this regularized point can result in larger estimation errors than simply using a zero initialization of the parameters \citenoh{manek2022pitfalls}.

Over-parameterization is another approach for ensuring the existence of a TD fixed point \citenoh{xiao2021understanding,thomas2022role}. 
In the offline setting, an over-parameterized model is defined to be one that has more parameters than the number of distinct data points in the dataset. 
Unfortunately, even over-parameterized TD still suffers from the deadly triad; in Figure \ref{fig:baird} below we observe that over-parameterized linear TD still diverges on Baird's counterexample \citenoh{baird1995residual}. 

Our paper establishes the convergence condition for TD with offline data when both a target network and an over-parameterized linear model are incorporated, as shown in Section \ref{sec:MRP}. 
Our condition is naturally satisfied for expected updates using the entire state-action space, ensuring guaranteed convergence.
Empirically, we demonstrate on Baird's counterexample that the over-parameterized target TD converges faster than other existing solutions to the deadly triad, such as residual minimization (RM) or gradient TD methods, while using less memory than convergent methods like LSTD.
Our result provides theoretical support for the empirical success of target networks and represents the first demonstration of a practical algorithm that is provably convergent and capable of high-quality empirical performance.

Importantly, over-parameterization ensures that the fixed point remains independent of the state collection distribution. 
Therefore, state distribution correction is not needed to approximate the value function of a target policy, which grants flexibility in collecting offline data from multiple state distributions. 
This property also means that the high variance and bias associated with state distribution corrections \citenoh{liu2018breaking} can be avoided, which has been a longstanding concern in the field. 
We show that the resulting fixed point, given full state coverage and accurate dynamics, approximates the target value function with an error that can be bounded by the distance between the best linearly represented value estimate and the true Q-value, $\frac{2}{1-\gamma} \inf_{\theta} \lVert \Phi \theta - q_{\pi} \rVert_{\infty}$, similar to the under-parameterized on-policy TD fixed point. 

Additionally, our convergence condition holds for learning with a batch of trajectory data collected from episodic MDPs. 
This result can be extended to truncated trajectory data with minor modifications, as explained in Section \ref{sec:NIS}. 
To compute the temporal difference error under the target policy, we consider two viable approaches: the first involves sampling the next action from the target policy, as detailed in Section \ref{sec:MRP}, while the second, introduced in Section \ref{sec:NIS}, adopts per-step normalized importance sampling (NIS) correction of action choices (not the state distribution) \citenoh{hesterberg1995weighted}. The latter method offers the advantage of converging with trajectory data under behaviour policies without making assumptions about the task or dataset. Consequently, we assert that \emph{the deadly triad issue can be fully resolved through the introduction of over-parameterized target TD with NIS correction over trajectory data}. The value estimation errors of these two approaches are compared empirically in a simple Four Room task in Section \ref{sec:NIS}. 

Finally, we extend the results to the offline control case. Q-learning \citenoh{watkins1992q} is a control algorithm based on temporal difference learning that also suffers from the deadly triad. 
Here we show that over-parameterized target Q-learning with offline data is also provably convergent.

\section{Background} \label{sec:background}

\paragraph{Notation} 
We let $\Delta(\cX)$ denote the set of probability distributions over a finite set $\cX$. 
Let $\RR$ denote the set of real numbers, and $\mathds{1}$ be the indicator function.
For a matrix $A\in\RR^{n\times m}$, we let $A^\dagger$ denote its Moore-Penrose pseudoinverse and $\rho(A)$ denote its spectral radius. 
Finally, 
we let $\text{diag}(x)\in\RR^{d\times d}$ be a diagonal matrix whose diagonal elements are given by $x\in\RR^d$.

\paragraph{Markov Decision Process}
We consider finite Markov Decision Process (MDP) defined by $\mathcal{M}=\{\cS, \cA, P, r, \gamma\}$, 
where 
$\cS$ is a finite  state space,
$\cA$ is the action space, 
$r:\cS\times\cA\rightarrow \RR$ is the reward function bounded by one,
$P:\cS\times\cA\rightarrow \Delta(\cS)$ is the transition matrix, and
$\gamma < 1$ is the discount factor.   
The \emph{Q-value} represents the expected cumulative rewards starting from a state-action pair $(s,a)$ following a policy 
$\pi:\cS\rightarrow \Delta(\cA)$, defined as
\begin{align*}
q_\pi(s, a) = \EE_\pi\left[ \sum_{t=0}^\infty \gamma^t r(S_t, A_t) \Big| S_0=s, A_0=a\right]\, ,
\end{align*}
where we use $\EE_\pi$ to denote the expectation under the distribution induced by 
$\pi$ and the environment.  
The \emph{Bellman operator} under the policy $\pi$ on 
$q(s,a)$ is defined as
\begin{equation*}
\mathcal{T}_{\pi}q(s,a) = r(s,a) +\gamma \sum_{s', a'\in\cS \times\cA} P_{\pi}(s', a'|s,a)q(s',a'), 
\end{equation*}
with the state-action transition distribution under $\pi$ defined as $P_{\pi}(s',a'|s,a) = P(s'|s,a) \pi(a'|s') $.  
We represent functions as vectors to enable vector-space operations: the value function and reward function are denoted by $q_\pi, r \in \RR^{|\cS||\cA|}$, and the transition function by $P_\pi \in \RR^{|\cS||\cA|\times |\cS||\cA|}$.  
Then, we define the \emph{Bellman operator} on any vector $q$ as 
\begin{align*}
\mathcal{T}_{\pi}q = r + \gamma P_\pi q\ .    
\end{align*}
It is known that the value function satisfies the \emph{Bellman equation} $q_\pi = \mathcal{T}_{\pi} q_{\pi}$. 

\paragraph{Linear Function Approximation}
In this work, we focus on linear function approximation, $q_\pi(s, a) \approx \phi(s, a)^\top\theta$, where $\theta\in\RR^d$ is a parameter vector, and $\phi:\cS\times\cA\rightarrow\RR^d$ maps a given state-action pair to a $d$-dimensional feature vector $\phi(s, a)\in\RR^d$. 
We denote $\Phi\in\RR^{|\cS||\cA|\times d}$ as the feature matrix, where each row corresponds to the feature vector of a particular state-action pair $(s, a)$. This matrix form allows us to write the value function approximation as $q_\pi \approx \Phi \theta$ for some parameter $\theta$. 
Finally, we assume that $\Phi$ is full rank, meaning there are no redundant features.  

\paragraph{Offline Value Prediction}

We consider \emph{offline value prediction}, that is, learning to predict the value of a target policy given a prior collected dataset $\cD$ that consists of transition data 
$\{(s_i, a_i, r_i, s_i')\}_{i=1}^{n}$. 
Let $\lambda\in \Delta(\cS\times\cA) $  be an arbitrary data collection distribution; the transition data is collected by first sampling a state-action pair $s_i, a_i$ from $\lambda$, then receiving the reward $r_i = r(s_i, a_i)$ and next state $s_i\sim P(\cdot | s_i, a_i)$ from the environment. 
The problem is known as {on-policy} if the data collection distribution $\lambda$ is the stationary distribution of $\pi$ and {off-policy}, otherwise \citenoh{sutton2018reinforcement}. 
Our results can also be extended for \emph{offline policy optimization}, aiming to extract a good control policy from the offline data. 

For clarity, we introduce the following additional notation.  
Given offline data $\cD$, let $n(s,a) = \sum_{i=1}^n \mathds{1}[s_i=s, a_i=a]$ be the count of a state-action $(s,a)$ observed in the data, and $\hat{\lambda}(s,a) = n(s,a) / n$ be the empirical distribution of $(s,a)$.  
We let $\{(s_i, a_i)\}_{i=1}^k \subseteq \cS\times\cA$ denote the state-actions with $n(s_i, a_i)>0$, where $k=\sum_{s,a} \mathds{1}[n(s,a) > 0]$ represents the count of state-action pairs with observed outgoing transitions. 
Following these definitions, 
we can define a mask matrix $H\in\RR^{k\times |\cS| |\cA|}$, and the empirical distribution of observed data ${D}_k\in \RR^{k\times k}$ as:
  \begin{align}
    H = \begin{bmatrix}
          \boldsymbol{1}_{x_1}^\top\\
          \vdots \\
          \boldsymbol{1}_{x_k}^\top
         \end{bmatrix}\, ,
    \qquad
    {D}_k = \begin{bmatrix}
            \hat{\lambda}(x_1) & & \\
            & \ddots & \\
            & & \hat{\lambda}(x_k)
  \end{bmatrix}\, ,
  \end{align}
where $\boldsymbol{1}_{(s_i, a_i)}\in\{0, 1\}^{|\cS||\cA|}$ is an indicator vector such that $\phi(s_i, a_i)=  \Phi^\top \boldsymbol{1}_{(s_i, a_i)}$.  
    
To evaluate the value of a target policy $\pi$, we augment each transition $(s_i, a_i, r_i, s'_i)$ to $(s_i,a_i,r_i,s_i',a_i')$ by selecting an action $a_i' \sim \pi(\cdot|s_i')$.  
The empirical transition matrix between state-action pairs $\hat{P}_\pi \in \mathbb{R}^{ |\mathcal{S}||\mathcal{A}| \times |\mathcal{S}| |\cA|}$ can then be defined 
for all $s', a'$
as:  
\begin{equation}
    \hat{P}_\pi(s', a' | s,a) = \frac{\sum_{i=0}^n \mathds{1}[s_i=s, a_i=a, s_i'=s', a'_i = a']}{ n(s,a)},
\end{equation}
if $n(s,a)>0$;
otherwise, $\hat{P}_\pi(s', a' | s,a)=0$. 
 
Given these notations, we can then define the \emph{empirical mean squared Bellman error (EMSBE)} as:
\begin{align}
\text{EMSBE}{(\theta)} =\frac{1}{2} \left \Vert {R} + \gamma N \theta - M \theta \right \Vert_{D_k}^2,
\label{eq:emsbe}
\end{align}
where $M = H\Phi\in\RR^{k\times d}$ denotes the predecessor features observed in the offline data, $N = H\hat{P}_\pi \Phi \in \RR^{k\times d}$ gives the next state-action features under the empirical transition, and ${R} = H r\in\RR^{k}$ gives the rewards.

\paragraph{Over-parameterization} This work considers the over-parameterization setting, such that the function approximation applies linear features with dimension $d > k$, the support of the empirical data. 
This allows all of the Bellman consistency constraints to be satisfied on all transitions in the offline data, driving EMSBE exactly to zero. In the over-paramterized regime, the Moore-Penrose pseudoinverse of the feature matrix, denoted as $\Phi^\dagger$, equals $\Phi^\top (\Phi \Phi^\top)^{-1}$.

\section{Over-parameterized Target TD}\label{sec:MRP}

We first show that leveraging \emph{overparameterization}  and \emph{target network} can significantly stabilize temporal difference learning with function approximation. 
We will also use the Baird counterexample to illustrate the effectiveness of over-parameterized target TD  \citenoh{baird1995residual}. 

\subsection{Over-parameterized TD Learning}
\begin{figure*}[ht]
    \centering
    \includegraphics[width=0.99\linewidth]{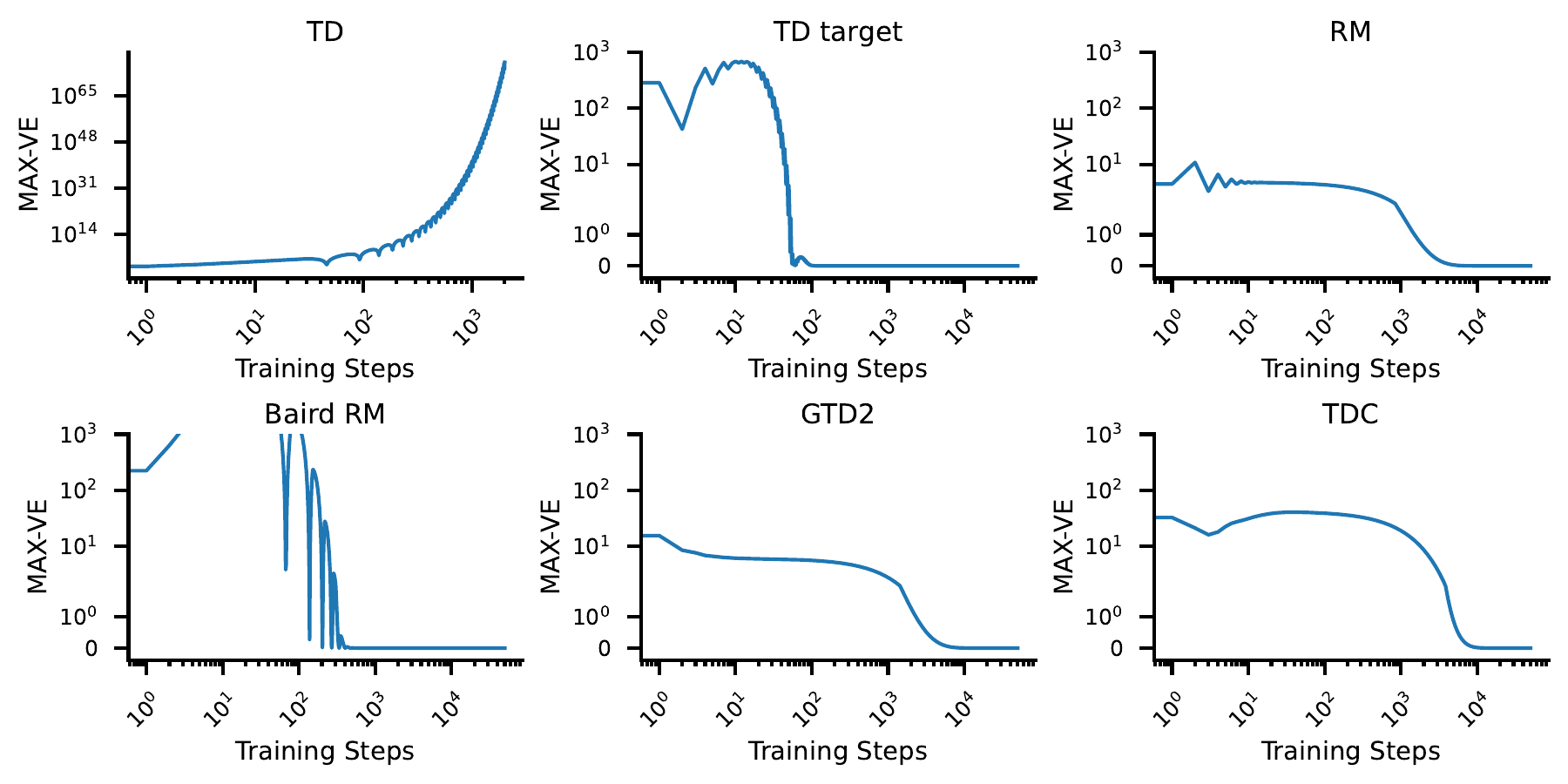}
    \vspace*{-3mm}
    \caption{On Baird counterexample, states are sampled from a uniform distribution, and there exists only one action at each state. The discount factor is set to be $\gamma=0.95$. We plot the maximal value prediction error among all states for OTD, OTTD, RM, and GTD algorithms. Other than OTD, the value errors converge to zero for the rest algorithms. OTTD avoids the divergence of TD and slow convergence rate of others.}
    \vspace*{-3mm}
    \label{fig:baird}
\end{figure*}

First, we briefly review the over-parameterized TD (OTD) algorithm for offline value prediction.  
OTD using the full batch of the dataset turns out to apply semi-gradients of EMSBE \eqref {eq:emsbe} to  update the parameter recursively,
\begin{align}
    \theta_{t+1} &= \theta_t + \frac{\eta}{n } \sum_{i=1}^n \delta_i \phi(s_i,a_i)\nonumber\\
    &= \theta_t - \eta M^\top D_k \left [M\theta_t - (R + \gamma  N \theta_{t}) \right].
    \label{eq:td}
\end{align}
where $\eta>0$ is the learning rate and $\delta_i = r_i + \gamma \phi(s_i’,a_i’)^\top \theta_t - \phi(s_i,a_i)^\top \theta_t$ is the TD error for each transition. 
\citetnoh{xiao2021understanding} analyzed the convergence properties of OTD, stated as follows.

\begin{proposition} 
For the over-parameterized regime $d>k$, 
if the following two conditions hold: 
\begin{itemize}
\setlength{\itemindent}{-1em}
\item $\rho(I - \eta  (M-\gamma N) M^{\top} D_k)<1$; 
\item $NM^{\dagger}$ has any sub-multiplicative norm smaller than or equal to one,
\end{itemize} 
then there exists a learning rate $\eta$ such that the parameter of OTD updates converges to
    \[\theta_{\mathrm{TD}}^* = M^{\dagger}(I-\gamma NM^{\dagger})^{-1}R,\]
    when the initial parameter of OTD equals zero.
    \label{prop:otd-converge}
\end{proposition}
It can be verified that the OTD fixed point $\theta_{\mathrm{TD}}^*$ is the minimum norm solution of EMSBE that lies in the span of $M$. 
Importantly, \cref{prop:otd-converge} characterizes an implicit algorithmic bias of OTD: it implicitly regularizes the solution toward a unique fixed point, even when  EMSBE admits infinitely many global minima. 
However, the convergence of OTD requires certain constraints on the features to guarantee convergence. 
In particular, it requires the matrix $I - \eta (M-\gamma N) M^{\top} D_k $ to have a spectral radius of less than one, which cannot be easily met on problems, including the Baird counterexample, and thus causes divergence, as we will show later. The failure of this condition is identified as a core factor behind the deadly triad issue, causing the update to be non-contractive \citenoh{sutton2016emphatic,fellows2023target}.
Furthermore, a sufficient condition would be to have orthonormal feature vectors and states showing up uniformly to satisfy the spectral radius property, which 
leaves no generalization space and creates an awkward tradeoff between instability and generalization. We did not find other sufficient conditions that can be easily checked.

\subsection{Over-parameterized Target TD}
Our first main contribution is to confirm theoretically 
that leveraging a target network significantly increases TD's stability with function approximation. 
Let $\theta_{\mathrm{targ}, t}\in\RR^d$ be the \emph{target parameter} at iteration $t$. 
We refer to the function approximation's original parameter $\theta_t$ as the \emph{student parameter}. 
Over-parameterized Target TD (OTTD) considers the following update rule using the entire offline dataset: 
\begin{equation}
    \theta_{t+1} = \theta_t - \eta M^{\top}D_k \left [M\theta_t - (R + \gamma  N \theta_{\mathrm{targ},m \lfloor \frac{t}{m} \rfloor}) \right].
    \label{eq:original_update_rule}
\end{equation}
That is, a target parameter $\theta_{\mathrm{targ}}$ is introduced to provide bootstrapping targets for TD updates. 
The target parameter is initialized with the student parameter $\theta_{\mathrm{targ}, 0} = \theta_0$, and is kept fixed for a window size $m$. 
Then, for every $m$ step, we update the target parameter by directly copying from the student parameter, 
$$\theta_{\mathrm{targ},(n+1)m} = \theta_{nm}.\footnote{That is, we are considering the hard target network update popularized by \citenoh{mnih2015human} instead of Polyak averaging. }$$
Our next result characterizes the convergence of OTTD. 
\begin{theorem}
For the over-parameterized regime $d>k$, 
given that the following condition holds:
\begin{itemize}
\setlength{\itemindent}{-1em}
\item $NM^{\dagger}$ has any sub-multiplicative norm smaller than or equal to one,
\end{itemize} 
 there exists a learning rate $\eta$ and an integer $\bar{m}$ such that for all update window sizes of the target parameters $m \ge \bar{m}$, the parameter of OTTD converges to 
    \[\theta_{\mathrm{TD}}^* = M^{\dagger}(I-\gamma NM^{\dagger})^{-1}R ,\]
when the initial point equals zero.
    \label{thm:target_TD_convergence}
\end{theorem}

\begin{remark}
The dependence on the initial point is detailed in Theorem \ref{thm:full_OTTD_initial_point} in \ref{A:convergence}.
The analytical form of $\Bar{m}$ depends on the specific norm constraint applied to $NM^{\dagger}$. For instance, when bounding the infinity norm, $\Bar{m} = 1+ \lceil \cfrac{\log(1-\gamma) - \log((1+\gamma)\sqrt{k})}{\log(1-\eta \lambda_{\mathrm{min}}(MM^TD_k))} \rceil$.
\end{remark}

\cref{thm:target_TD_convergence} illustrates the efficacy of incorporating a target network in stabilizing bootstrapping with function approximation. This is apparent in the convergence of OTTD to the TD fixed point, eliminating the condition on the spectral radius of $I - \eta (M-\gamma N) M^{\top} D_k$ to be bounded by one.  
Central to the convergence of OTTD is the role of $NM^{\dagger}$, which represents the projection coefficient of each row of $N$ onto the row space of $M$. When applying bootstrapping, values are extrapolated using $N M^{\dagger} M \theta$, based on these projected coefficients. When the infinity norm of  $NM^{\dagger}$ is bounded by one, we have $\lVert N M^{\dagger} M \theta \rVert_{\infty} \le \lVert M \theta \rVert_{\infty}$. Thus, the condition on the norm of $NM^{\dagger}$ prevents overestimation for bootstrapping values outside the current state-action set. 
Although OTTD still requires this condition for convergence, as we will show later in the paper, it can be resolved when the offline dataset consists of trajectory data.

\subsection{Special Case Analysis of Expected Updates}
\label{sec:full-update}

The benefit of incorporating a target network is fully presented when considering expected updates with an off-policy data distribution that covers the entire state-action space. Off-policy learning means that we do not constrain the data distribution $\lambda$ to be the stationary distribution under a target policy. In this context, OTTD naturally converges with a proven guarantee. In contrast, using only one augmentation—either a target network or an over-parameterized model—can still result in divergence.

The expected update rules for TD and TD with a target network (target TD) are the same for under-parameterized or over-parameterized models and are outlined below:
\begin{itemize}
\item TD expected update: 
\begin{align*}
    \theta_{t+1} = \theta_t - \eta \Phi^\top D \left [\Phi\theta_t - (R + \gamma  P_\pi \Phi \theta_{t}) \right].
\end{align*}
\item Target TD expected update: 
\begin{align*}
    \theta_{t+1} = \theta_t - \eta \Phi^\top D \left [\Phi \theta_t - (R + \gamma  P_\pi \Phi \theta_{\mathrm{targ},m \lfloor \frac{t}{m} \rfloor}) \right].
\end{align*}
\end{itemize}

\begin{figure}
    \centering
\vspace{-1cm}
  \begin{tikzpicture}
  \node[circle, draw, line width=0.4mm, minimum size=1cm] (theta) at (0,0) {$\theta$};
  \node[circle, draw, line width=0.4mm, minimum size=1cm] (2theta) at (2,0) {$2\theta$};

  \draw[->, line width=0.3mm] (theta) -- (2theta) node[midway, above] {0};
  \draw[->, line width=0.3mm, loop, out=45, in=-45, looseness=8] (2theta) to  node[midway, right] {0} (2theta);
\end{tikzpicture}    
\vspace{-1.25cm}
    \caption{In this example, each state has exactly one action, and rewards are labeled next to the transitions. The value functions are parameterized by a scalar parameter $\theta$, and the features are shown in the graph. This counterexample demonstrates a task with fixed transition probabilities and rewards where our convergence condition is satisfied. However, the conditions for under-parameterized target TD fail for certain data distributions.}
    \label{fig:two-state}
\end{figure}
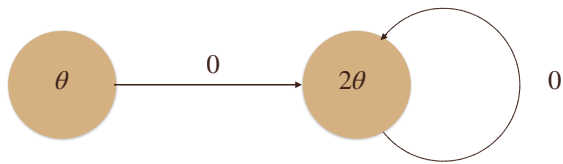

Several works \citenoh{asadi2023td,fellows2023target} have analyzed the convergence conditions for under-parameterized target TD. However, their condition, which requires $(\Phi^\top D \Phi)^{-1}\Phi^\top D \gamma P \Phi$ to have a spectral radius or norm of less than one as listed in \ref{A:assump}, cannot be met for all off-policy data distributions. This limitation is demonstrated by a Two-state counterexample in Figure \ref{fig:two-state}, where certain data distributions result in the fixed point of under-parameterized TD not existing, causing the convergence condition to fail.

In the counterexample, the feature matrix of these two states equals to $\Phi = \begin{pmatrix}
    1\\
    2
\end{pmatrix}$. The learning is off-policy if the state distribution differs from the stationary distribution, which concentrates on the right state with self-loop. For any discount factor $\gamma > 0.5$, the off-policy state distribution $(\frac{4\gamma-4}{2 \gamma -3},\frac{1-2\gamma}{2 \gamma -3})$ causes the fixed point to be $(\Phi^\top D (I -\gamma P) \Phi)^{-1} \Phi^\top D R = \frac{0}{0}$, which does not exist. Consequently, the learning of TD, with or without a target network, is stuck at any initialization. This distribution and any others weighing even less on the right state transition break the required condition for the convergence proof in Corollary 2 of Asadi et al.\ (2023) and the non-asymptotic behaviour analysis in Corollary 3.1 of Fellows et al.\ (2023).

\begin{table}
\begin{center}
\vspace{-3mm}
\caption{The table shows a metric for the convergence rate. A higher value of this metric indicates a slower convergence rate, and a larger-than-one value represents the divergence of the algorithm.}
\vspace{2mm}
\begin{tabular}{ c |c |c }
 Algorithm & Convergence? & $\lim_{t\to \infty} \lVert C \rVert^{\frac{1}{t}}$ \\
 \hline
 TD & No & $1.12$ \\ 
 \hline
 Target TD & Yes & $1- 3.8\mathrm{e}{-3}$\\
 \hline
 RM & Yes & $1- 1.9\mathrm{e}{-5}$ \\ 
 \hline
 GTD2 & Yes & $1- 4.5\mathrm{e}{-6}$\\
\end{tabular}
\end{center}
\label{tab:conv_metrix}
\end{table}
\vspace{-1mm}

Adding over-parameterization alone does not ensure convergence either. As proven in Proposition \ref{prop:otd-converge}, the convergence of OTD still requires that $\rho(I - \eta \Phi^T D (\Phi-\gamma P_\pi \Phi))<1$, which can be easily violated.
Baird's counterexample \citenoh{baird1995residual} illustrates this expected update with all states observed uniformly. Since each state has only one action, actions must adhere to the target policy, as requested by the data collection procedure in Section \ref{sec:background}. Yet, states are sampled from an off-stationary distribution, giving the challenges of off-policy learning. The features are intentionally over-parameterized, having more dimensions than the total number of states. Hyperparameters' choices are in \ref{A:exper}.
As shown in Figure \ref{fig:baird}, OTD diverges, while OTTD successfully prevents the divergence observed in standard TD.

In summary, over-parameterization fundamentally eliminates the model's dependency on the data generation distribution, and a target network removes a core factor behind the deadly triad issue, causing the update to be non-contractive \citenoh{sutton2016emphatic,fellows2023target}. Therefore, the combination of these two simple augmentations resolves the deadly triad, as stated in Proposition \ref{prop:OTTD_expected_updates_convergence}.

\begin{proposition} 
For the over-parameterized regime $d>k$, there always exists a learning rate $\eta$ and an integer $\bar{m}$ such that for all update window sizes of the target parameters $m \ge \bar{m}$, the parameter of OTTD converges to 
    \[\theta_{\mathrm{TD}}^* = \Phi^{\dagger}(I-\gamma P_{\pi})^{-1}R + (I - \Phi^{\dagger}\Phi)\theta_0,\]
    where $\theta_0$ is the initial parameter of OTTD.
    \label{prop:OTTD_expected_updates_convergence}
\end{proposition}

We also compare OTTD with other standard algorithms empirically on the Baird's counterexample, including Residual Minimization  (RM), Baird RM \citenoh{baird1995residual}, Gradient TD2 (GTD2) \citenoh{maei2011gradient}, and TDC \citenoh{sutton2009fast}. 
Figure \ref{fig:baird} indicates that OTTD also mitigates the slow convergence rates of alternative algorithms. 
The update rules of these algorithms can be expressed through the equation $\theta_{t} - \theta^* = C^{t} (\theta_0-\theta^*)$. To gauge the convergence speed of these algorithms, Schoknecht and Merke (2003) \nocitenoh{schoknecht2003td} introduced a metric defined as $\lim_{t\to \infty} \lVert C_{\text{target}} \rVert^{\frac{1}{t}}$. A higher value of this metric indicates a slower convergence rate. The convergence metrics, presented in Table \ref{tab:conv_metrix}, reveal that the metric exceeds one for TD, leading to divergence. On the other hand, the metrics for RM and GTD2 hover too close to one, resulting in slower convergence rates.

\subsection{Value Prediction Error Bound}
We present a worst-case value prediction error bound of OTTD across the entire state-action space.

\begin{theorem}
    Given $\lVert N M^{\dagger} \rVert_{\infty}<1$, 
    with probability at least $1-\delta$, the fixed point of OTTD $\theta^*_{TD}$ learnt with the dataset $\mathcal{D}$ gives the following value prediction error bound
    \begin{align}
       \lVert \Phi \theta_{\mathrm{TD}}^*-q_{\pi} \rVert_{\infty} \le \epsilon_{stat}+\epsilon_{projection}+\epsilon_{approx},
    \end{align}
where error terms are defined as follows:
{\setstretch{0.64}
\begin{itemize}
    \setlength{\itemindent}{-1em}
    \item Statistical error \(\epsilon_{stat}\) equals
    \[\epsilon_{stat} = \frac{\lVert \Phi M^{\dagger} \rVert_{\infty}  }{(1-\gamma)^2}\sqrt{\frac{\log(\frac{2k|\mathcal{A}|}{\delta})}{\min_{(s,a)}n(s,a)}}.\]
    \item Projection error \(\epsilon_{projection}\) is defined as
    \[\epsilon_{projection} = \frac{\lVert \Phi M^{\dagger} \rVert_{\infty}}{1-\gamma} \lVert \Phi(I- M^{\dagger}M) \theta^* \rVert_{\infty}.\]
    \item Approximation error \(\epsilon_{approx}\) is given by
    \[\epsilon_{approx} = \frac{2 \lVert \Phi M^{\dagger} \rVert_{\infty}}{1-\gamma} \lVert \Phi \theta^* -q_{\pi} \rVert_{\infty}.\]
\end{itemize} }
\vspace*{-2mm}
    \label{thm:target_td_gen_error}
\end{theorem}

Here, \(\theta^* = \arg\max_{\theta} \lVert \Phi \theta - q_{\pi} \rVert_{\infty}\) is the optimal parameter minimizing the difference between \(\Phi \theta\) and \(q_{\pi}\), $\epsilon_{stat}$ counts in the estimation error of the MDP dynamics by the dataset, and $\epsilon_{projection}$ accounts for insufficient coverage of the dataset. Notably, in overparameterization, information perpendicular to the row space of $M$, the space of data features, decides the size of errors instead of distribution shift ratios. The proof is given in \ref{A:error}.

Next, we narrow our focus on the expected update discussed in \cref{sec:full-update}. 
This scenario, characterized by $M=\Phi$ and $N = P_\pi \Phi$, remains off-policy. The fixed point of OTTD $\theta^*_{\mathrm{TD}} = \Phi^{\dagger}(I-\gamma P_{\pi})^{-1} r$, gives the following error bound.
\begin{corollary}
    For the optimal fixed point $\theta^*_{\mathrm{TD}} = \Phi^{\dagger}(I-\gamma P_{\pi})^{-1} R$, the approximation error is bounded as
    \begin{align}
       \lVert \Phi \theta^*_{\mathrm{TD}}-q_{\pi} \rVert_{\infty} \le \frac{2}{1-\gamma} \inf_{\theta} \lVert \Phi \theta -q_{\pi} \rVert_{\infty}.
    \end{align}
    \label{thm:gen_error_true_MDP}
\end{corollary}

The value prediction error bound of off-policy OTTD closely aligns with the on-policy results obtained using under-parameterized linear models. 
This significance becomes evident when considering the stringent requirement in on-policy learning, where data must be sampled from the stationary distribution $d_{\pi}$ of the target policy $\pi$. This key observation underscores the remarkable ability of over-parameterized models to learn irrespective of the underlying data distribution $\lambda$. 
To elaborate, Tsitsiklis and Van Roy (1997) have shown that on-policy TD fixed point $\theta_{\mathrm{under}}^*$ in the under-parameterized setting satisfies that $\lVert \Phi \theta_{\mathrm{under}}^*-q_{\pi} \rVert_{D_{\pi}} \le \frac{1}{1-\gamma} \inf_{\theta}\lVert \Phi \theta_{\mathrm{under}}^*-q_{\pi} \rVert_{D_{\pi}}$, where $D_{\pi} = \text{diag}(d_{\pi})$. It aligns with the error bound given here, differing only in a constant and a norm.
\vspace{-2mm}
\section{Learning with Normalized IS Correction}\label{sec:NIS}

Next, we show that using offline trajectory data can remove the remaining convergence condition of OTTD. Establishing the convergence without relying on specific assumptions about the tasks or features signifies that the deadly triad issue is resolved.

In this section we leverage trajectory data, where the state-action pairs to be bootstrapped are also trained, except for the last states. Thus, the condition of limiting overestimation on out-of-dataset value estimates is no longer required.
The dataset $\mathcal{D}= \{\tau_j \}_{j=1}^{n'}$ consists of $n'$ trajectories.
Each trajectory \(\tau_j\) is a sequence of state-action-reward tuples sampled under a behavior policy \(\mu\), defined as \(\tau_j = \{(s_t^j, a_t^j, r_t^j, s_{t+1}^j)\}_{t=0}^{T_j-1}\), with \(T_j\) indicating the length of the trajectory. In this context, \(a_t^j \sim \mu(\cdot | s_t^j)\) and \(s_{t+1}^j \sim P(\cdot | s_t^j, a_t^j)\) is generated by the MDP.

We handle the final states of each collected trajectory \(\{s_{T_j}^j\}_{j=1}^{n'}\), by implementing a looping mechanism that sets these states to transition back to themselves with zero rewards. Specifically, an additional transition \((s_{T_j}^j, a_{T_j}^j, 0, s_{T_j}^j)\) is appended to each trajectory \(\tau_j\), with  \(a_{T_j}^j \sim \mu(\cdot | s_{T_j}^j )\). 
This setup aligns with episodic tasks, where each episode ends at a terminal state with zero-reward self-loop transitions \citenoh{sutton2018reinforcement}. 
However, it does introduce some challenges in continuing tasks, leading to additional errors in value predictions: the details of these prediction errors are  discussed in \cref{sec:error bound}. Consequently, each trajectory can be decomposed into transitions $\{(s_i,a_i,r_i,s_i',a_i')\}_{i=1}^n$.

\begin{figure*}[ht]
    \centering
    \hspace*{-3mm}
    \vspace*{-3mm}
    \includegraphics[width=1\linewidth,height=0.27\linewidth]{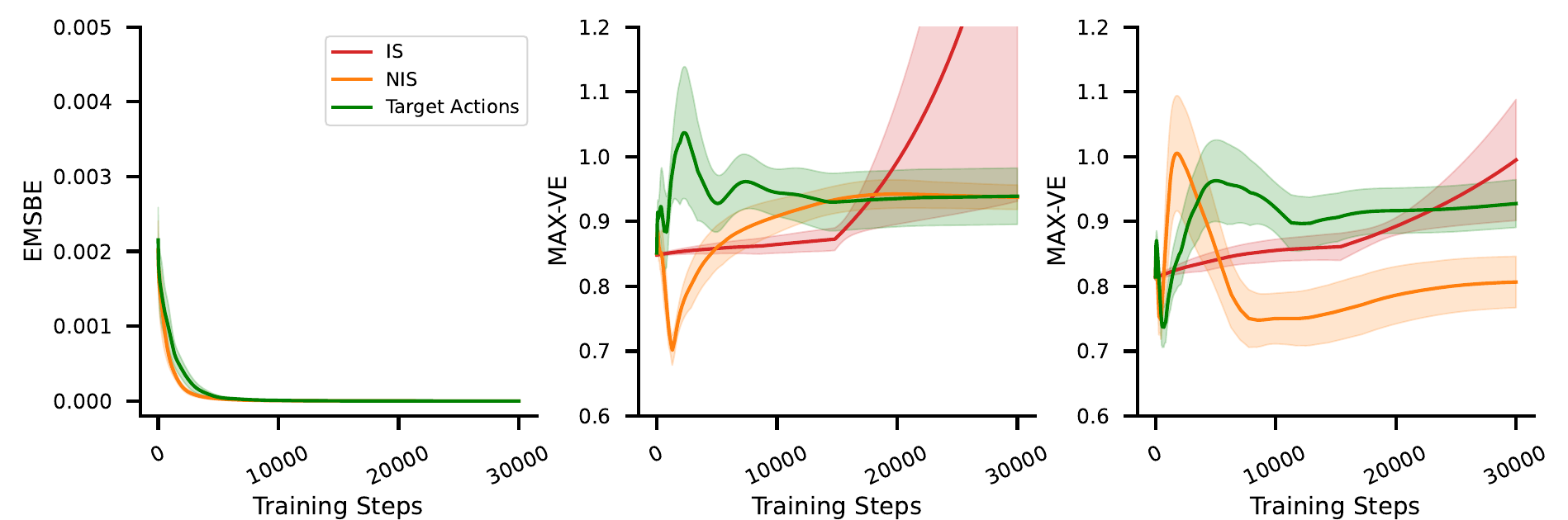}
    \vspace*{1mm}
    \caption{On Four Room, data are sampled as trajectories under the random policy, while the target policy is given by a human player. The left sub-figure shows the training error, EMSBE. The middle and the right sub-figures show the infinity norm of value errors, that is, $\lVert \Phi \theta_{\mathrm{TD}}^*-q_{\pi} \rVert_{\infty}$. Here, the right sub-figure uses a larger dataset, and all results are averaged over $10$ random seeds. With off-policy data, per-step normalized IS can correct the action distribution and behave similarly to sampling actions from the target policy. Due to less variance, normalized IS avoids divergence of IS correction.}
    \vspace*{-4mm}
    \label{fig:room_IS}
\end{figure*}

Given this setup, we apply importance sampling (IS) corrections to align off-policy data distributions with the target policy \(\pi\). 
For each $(s_i',a_i')$ the corresponding IS ratio for fixing the next action's distribution as $\rho(a_i' | s_i') = {\pi(a_i' |s_i' )} / {\mu(a_i' | s_i')}$.  
This per-step action distribution correction suffices without any state distribution correction. The state-action transitions $\hat{P}_\pi(s', a' | s,a)$ can be estimated as
\begin{equation}
    \frac{\sum_{i=0}^n \rho(a'|s')\mathds{1}[s_i=s,a_i=a,s_i'=s',a_i'=a']}{n(s,a)},
\end{equation}
if \(n(s,a)>0\); otherwise, zero. However, the high variance introduced by the IS ratio can cause bootstrapping on extremely high values and lead to instability in learning, as illustrated in Figure \ref{fig:room_IS}. 
We thus leverage the Normalized Importance Sampling (NIS) correction \citenoh{hesterberg1995weighted,kuzborskij2021confident} to reduce the variance of the update.

\vspace{-1mm}
\subsection{Normalized Importance Sampling}

When using NIS to approximate the transition probability under the target policy $\pi$ from the state-action pair $(s,a)$, the sum of IS ratios for transitions from $(s,a)$ is used as the normalization term instead of the count $n(s,a)$. More specifically, each element of the estimated transition matrix, denoted as $\hat{P}_{\pi, \mathrm{NIS}}(s', a'|s,a)$ for $n(s,a)>0$, is defined as
\begin{equation}
    \frac{\sum_{i=0}^n \rho(a'|s')\mathds{1}[s_i=s, a_i=a, s_i'=s', a_i'=a']}{\sum_{i=1}^n \sum_{\Tilde{s},\Tilde{a}}\rho(\Tilde{a}|\Tilde{s})\mathds{1}[s_i=s, a_i=a, s_i'=\Tilde{s}, a_i'=\Tilde{a}]};
\end{equation}
otherwise, is set to zero.
The numerator summarizes the corrected occurrences of transitions into $(s',a')$ from the state-action $(s,a)$, and the denominator represents the total corrected occurrences of the state-action $(s,a)$.

The following proposition illustrates that our transition estimator is consistent in the absence of artificially added loop transitions. The proof is given in \ref{A:NIS}.

\begin{proposition}
    When behaviour policies cover the support of the target policy, the transition probability estimator $\hat{P}_{\pi}(s',a'|s,a)$ is consistent for all state-action $(s,a)$ without additional loop transitions, that is the estimator tends to $P_{\pi}(s',a'|s,a)$ almost surely as $n(s,a) \to \infty$. Here, $n(s,a)$ is the counts of the current state-action pair $(s,a)$.
\end{proposition}

Let $N_{\mathrm{NIS}}  = H \hat{P}_{\pi, \mathrm{NIS}} \Phi$ be the next state-action feature matrix under the NIS transition estimate $\hat{P}_{\pi,\mathrm{NIS}}$. 

The algorithm with NIS correction is realized by weighting each transition proportional to its IS ratio $\rho(a_i'|s_i')$ for the next action. It turns out that the update rule is equivalent to taking a semi-gradient of the weighted Bellman error and is given by
\begin{align}
\theta_{t+1} &= \theta_t + \frac{1}{\sum_{i=1}^n \rho(a_i' | s_i')} \eta \sum_{i=1}^n  \rho(a_i' | s_i') \delta_i \phi(s_i,a_i)\nonumber\\
&= \theta_t + \eta M^\top D_{k,\mathrm{NIS}} (R+\gamma \hat{P}_{\pi,\mathrm{NIS}} M \theta_{\mathrm{targ},t} - M \theta_t).
\end{align}
where $\delta_i$ is the TD error for each transition and the diagonal matrix $D_{k,\mathrm{NIS}}$ represents the weighted empirical distribution by the importance sampling ratios. Though the empirical distribution is modified, it does not influence the fixed point, as shown in the following theorem.

With modifications to trajectory data and the NIS correction, the condition on the matrix \(N_{\pi, \mathrm{NIS}} M^{\dagger}\) is naturally met. This matrix becomes equivalent to the non-zero square matrix in \(\hat{P}_{\pi, \mathrm{NIS}}\) and is stochastic, with its infinity norm equal to one. 
Therefore, the algorithm OTTD with NIS correction effectively addresses the deadly triad for off-policy tasks with trajectory data, as elaborated in the following theorem.

\begin{theorem}
    For the over-parameterized regime $d>k$, given a batch of trajectories, there exists a learning rate $\eta$ and integer $\bar{m}$ such that for all update window sizes for target parameters $m \ge \bar{m}$, the 
    OTTD update converges to 
    \[\theta_{\mathrm{TD}, \mathrm{NIS}}^* = M^{\dagger}(I-\gamma N_{\mathrm{NIS}}M^{\dagger} )^{-1}R + (I - M^{\dagger}M)\theta_0,\]
    where $\theta_0$ is the initial point.
    \label{thm:NIS_convergence}
\end{theorem}

\vspace{-1mm}
\subsection{Value Prediction Error Bound}\label{sec:error bound}
Our analysis first addresses 
value prediction errors in episodic tasks where trajectories end in terminal states. 
In these scenarios, loop transitions do not impact the error since terminal states inherently hold zero value. 
The primary findings of this analysis are detailed in the following corollary. The only distinction from the bound presented in Section \ref{sec:MRP} is the modification of the statistical error $\epsilon_{stat}$. It is adjusted due to the new transition probability estimator with NIS corrections. 

\begin{corollary}
Given a dataset $\mathcal{D}$ of episodic trajectory data collected under a behaviour policy $\mu$, 
when the following condition holds
\begin{itemize}
    \setlength{\itemindent}{-1em}
    \item $\mu$ covers the support of the target policy $\pi$, 
\end{itemize} 
with probability at least $1-\delta$, the fixed point of OTTD with NIS correction $\theta^*_{\mathrm{TD}}$ gives the following value prediction error bound
\begin{align}
    \lVert \Phi \theta_{\mathrm{TD}}^*-q_{\pi} \rVert_{\infty} \le \epsilon_{stat}'+\epsilon_{projection}+\epsilon_{approx},
\end{align}
where the projection error $\epsilon_{projection}$ and the approximation error $\epsilon_{approx}$ are the same as in Theorem \ref{thm:target_td_gen_error}.

The statistical error $\epsilon_{stat}'$ is defined as:
$$\epsilon_{stat}' = \frac{ \lVert \Phi M^{\dagger} \rVert_{\infty} \rho_M \max\{\rho_M-1,1\}}{(1-\gamma)^2 \sqrt{\min_{(s,a)}n(s,a)}}\log(\frac{4k|\mathcal{A}|}{\delta}),$$ 

with $\rho_M:=\cfrac{\max_{(s',a') \in \mathcal{D}} \rho(a'|s')}{\min_{(s',a') \in \mathcal{D}} \rho(a'|s')} $.
\vspace*{-2mm}
\end{corollary}

For truncated trajectories from continuing tasks, however, the integration of loop transitions gives samples that do not follow the MDP and thus introduces extra errors in estimating the reward and transition matrix. These discrepancies prevent establishing a meaningful bound using the infinity norm. Instead, we provide a bound based on mean squared error, elaborated in \ref{A:cont_error}.

\vspace{-1mm}
\subsection{Empirical Result}
In this section, we empirically analyze the value prediction errors in an episodic Four Room task using offline data from trajectories under a random behaviour policy. The target policy is chosen by a human player. 
The dataset only covers part of the entire state-action space, so generalization is required. Further details are provided in \ref{A:exper}. 

As shown in Figure \ref{fig:room_IS}, the IS correction, marked in red, can bootstrap on large values due to its high variance, which causes the condition to break, leading to divergence. By contrast, the other two OTTD methods successfully show zero EMSBE, 
as depicted in the left sub-figure. The first method, in green and detailed in Section \ref{sec:MRP}, selects the next action from the target policy, 
$a_i' \sim \pi(\cdot|s_i')$. The second method, in orange and covered in Section 4, leverages per-step normalized importance sampling (NIS) correction. 
The comparison of the infinity norm of value errors, $\lVert \Phi \theta_{\mathrm{TD}}^*-q_{\pi} \rVert_{\infty}$, between these two methods is illustrated in the middle and right sub-figures of Figure \ref{fig:room_IS}, with the latter sub-figure utilizing a larger dataset. These two methods exhibit equal performance, except 
as the dataset size increases, the error associated with NIS correction marginally decreases, a trend not observed with the target action sampling approach.

\section{Offline Control with Over-parameterized Target Q-learning}
\looseness=-1 The previous convergence result can be extended to Q-learning, a widely adopted TD algorithm for learning optimal policies. Q-learning often bootstraps on unobserved actions with the highest value estimates. However, we aim to avoid extrapolation beyond the dataset, thereby circumventing additional assumptions on unobserved data. To address this, we adapt  Q-learning  by limiting the $\arg\max$ operation to be over actions seen in the dataset, which is a common technique in offline learning \citenoh{kostrikov2021offline,xiao2022sample}. 
To develop the modified algorithm,
first define feature matrices $\Phi_i$ to present features for the state $s_i \in \mathcal{S}$ with only seen actions. If this state does not show up in the dataset, set the matrix as a vector of zeros.

Similar to OTTD, the target parameter at iteration $nm$, for $n=0,1,\cdots$, copies the student parameter, $\theta_{\mathrm{targ},nm} = \theta_{nm}$, and is kept fixed for $m$ steps.  
At each iteration $t$, over-parameterized target Q-learning (OTQ) first computes values for bootstrapping using the target parameter
\vspace{-.5mm}
\begin{align*}
    &q_t(s,a) =\phi(s,a)^{\top}\theta_{\mathrm{targ},m \lfloor \frac{t}{m} \rfloor},\\
    &y_t(s,a) = R(s,a)+ \gamma \sum_{s'} \hat{P}(s'|s,a)\max_{\text{\normalfont seen } a'} q_t(s',a'), 
\end{align*}
where we denote $\hat{P} \in \mathbb{R}^{k \times |\mathcal{S}|}$ the empirical transition matrix.
Then, the student parameter is updated as
\begin{align}
    \theta_{t+1} &= \theta_t - \eta \sum_{s,a} \hat{\lambda}(s,a) \phi(s,a) (\phi(s,a)^{\top}\theta_t -y_t(s,a) ) \nonumber\\
    &=\theta_t - \eta M^\top D_k (M \theta_t - R - \gamma \hat{P} N_t),
\end{align}
where $i$-th element of $N_t \in \mathbb{R}^{|\mathcal{S}|}$ equals $\lVert \Phi_i \theta_{\mathrm{targ},m \lfloor \frac{t}{m} \rfloor} \rVert_{\infty}$.

\begin{theorem}
    For the over-parameterized regime $d>k$, 
    given that the following condition holds
    \vspace{-2.5mm}
    \begin{itemize}
    \setlength{\itemindent}{-1em}
    \item $ \lVert \Phi_i^{\top} M^{\dagger} \rVert_{\infty}<\frac{1}{\gamma}$, for $i=1,\cdots,|\mathcal{S}|$,
    \end{itemize} 
    \vspace{-2.5mm}
    there exists  an 
    $\bar{m}$ such that for all update window sizes of the target parameters $m \ge \bar{m}$, the OTQ update  converges to $\theta^* = M^{\dagger}\hat{q}^*+(I - M^{\dagger}M) \theta_0$, where $\hat{q}^* \in \mathbb{R}^k$ satisfies
        \[\hat{q}^* = R + \gamma  H \hat{P} \begin{pmatrix}
    \lVert \Phi_1 M^{\dagger} \hat{q}^* \rVert_{\infty} \\
    \lVert \Phi_2 M^{\dagger} \hat{q}^* \rVert_{\infty} \\
    \cdots \\
    \lVert \Phi_{|\mathcal{S}|} M^{\dagger} \hat{q}^*\rVert_{\infty} \\
    \end{pmatrix},\]
        and $\theta_0$ is the initial point.
\end{theorem}
\vspace{-2mm}
\looseness=-1 Proof for results in this section are presented in \ref{A:Q}. 
Here, the $\arg\max$ operator is expressed by the maximum norm. The optimal Q-values for the dataset, $\hat{q}^*$, are defined only for state-action pairs in $\{(s_i,a_i)\}_{i=1}^k$ with outgoing transitions. This value may not be defined for other state-action pairs with only incoming transitions, whose values are needed for bootstrapping and rely on extrapolation. Since the dataset may not describe any exiting transition, their optimal Q-values cannot be evaluated through bootstrapping but only projecting as $\Phi_{i} M^{\dagger}\hat{q}^*$. The term $\Phi_{i} M^{\dagger}$, for $i=1,\cdots,|\mathcal{S}|$, represents the projection coefficient of features onto the row space of $M$. The values for state-action pairs without outgoing transitions depend on extrapolating $\hat{q}^*$ in proportion to the projection coefficient. The condition on the norm of $\Phi_{i} M^{\dagger}$ further prevents overestimating those extrapolated values. This condition evidences that avoiding overestimating out-of-distribution actions for learning stability is important.

In the scenario where the dataset consists of episodic trajectory data, seen state-action pairs and their transitions $\hat{P}$ form a truncated empirical MDP. Also, all data used for bootstrapping is included in the training set. As long as we constrain maximization over seen actions, there is no extrapolation for bootstrapping values, and the convergence is established without further assumptions. Here, $\hat{q}^*$ is the optimal Q-value on this empirical MDP. We present this particular scenario in the following result.

\begin{corollary}
    For the over-parameterized regime $d>k$, given a batch of episodic trajectories,
    there exists an integer $\bar{m}$ such that for all update window sizes of the target parameters $m \ge \bar{m}$, the parameter of OTQ converges to $\theta^* = M^{\dagger}\hat{q}^*+ (I- M^{\dagger}M) \theta_0$, where
    \[\hat{q}^*(s,a) = R(s,a) + \gamma \sum_{s'} \hat{P}(s' | s,a) \max_{\text{\normalfont seen }a'}\hat{q}^*(s',a').\]
\end{corollary}
\vspace{-4mm}

\section{Related Work}
Most analyses on the convergence of TD are done in online settings. At each step, the parameters are updated by one transition coming in online, either as trajectories or sampled i.i.d.
TD with linear function approximation converges when data is sampled as trajectories under the target policy \citenoh{tsitsiklis1996analysis,dayan1992convergence} and adding a target network gives the same fixed point \citenoh{lee2019target}. 
But linear TD with off-stationary state distribution is not guaranteed to converge. This issue is called the deadly triad. 

When the fixed point of TD exists, gradient TD methods \citenoh{sutton2009fast} converge, but much more slowly than regular TD and are sensitive to hyperparameters \citenoh{maei2011gradient,sutton2008convergent,mahadevan2014proximal}. Several algorithms based on gradient TD have been proposed to prevent divergence and gain close-to-TD performance, but do not fully overcome the disadvantages and under-perform empirically 
\citenoh{ghiassian2020gradient,mahmood2017multi}. Regularized least square TD (LSTD) \citenoh{boyan1999least,lagoudakis2003least,kolter2009regularization,yu2009convergence} with L2 regularization converges to a regularized fixed point, but it stores a feature matrix of dimension $d\times d$, which is extreme in the over-parameterized setting. The convergence and generalization properties of over-parameterized TD have been discussed in \citenoh{xiao2021understanding,thomas2022role}. 

Several papers suggest that a target network may help TD overcome the deadly triad \citenoh{zhang2021breaking, chen2023target,asadi2023td,fellows2023target}. Some studies employ various modifications to analyze the target network, such as updates with value truncation or parameter projection. Moreover, additional assumptions are frequently required to establish convergence results, which cannot be met for all data distributions, even in expected updates over the entire state-action space.

Residual minimization (RM) \citenoh{baird1995residual} is known to converge with linear function approximation under any state distribution, and the convergence of over-parameterized RM is also confirmed \citenoh{xiao2021understanding}. However, RM typically converges slower than TD, as observed empirically \citenoh{gordon1995stable,van2011insights} and proven in the tabular setting \citenoh{schoknecht2003td}. Baird's residual algorithm \citenoh{baird1995residual} merges the parameter updates for TD and RM, offering enhanced stability but still a slower convergence rate than traditional TD learning. 

Other methods correct the data distribution by importance sampling \citenoh{precup2000eligibility,precup2001off,mahmood2014weighted,hesterberg1995weighted}. However, these approaches suffer from high variance when correcting the distributions of trajectories with products of IS ratios. Later papers work on estimating state distribution ratios to avoid the ratio product. However, these methods have not yet been adopted for practical algorithms since they are still suffering from more significant variances, biases, and computation requirements.\citenoh{sutton2016emphatic,hallak2017consistent,gelada2019off,nachum2019dualdice,nachum2019algaedice,yang2020off,zhang2020gradientdice,che2023correcting,he2023loosely}.

In contrast, offline RL, also known as batch RL, considers the setting where no online interactions in environments are allowed and often suffers from insufficient state space coverage and distribution shifts \citenoh{levine2020offline}. Avoiding overestimation for out-of-distribution action values stabilizes the learning \citenoh{fujimoto2018addressing,kumar2019stabilizing} and common techniques include constraining action selections \citenoh{kostrikov2021offline,hu2023sample,xiao2022sample}, adding pessimism \citenoh{jin2021pessimism} and limiting learned values directly \citenoh{kumar2020conservative}. Our modified over-parameterized target Q-learning also confines maximum action selection among seen actions in the dataset.

\section{Conclusion}
The susceptibility of temporal difference learning to divergence has been a longstanding challenge. Numerous algorithms and techniques have been explored to address this issue, but achieving a balance between stability and performance still needs to be achieved. Our paper first demonstrated that the combination of a target network and an over-parameterized model provided a principled solution to the challenges faced by TD in off-policy learning. While our study is currently confined to linear function approximation, it offers compelling evidence for convergence guarantees and the existence of a qualified fixed point. Extending these results to neural networks would be a crucial next step to understanding TD's empirical success with target networks.

\vspace{-2mm}
\section*{Impact Statement}
\vspace{-1mm}
This paper presents work whose goal is to advance the field of Machine Learning. There are many potential societal consequences of our work, none of which we feel must be specifically highlighted here.

\section*{Acknowledgements}
We sincerely appreciate Zechen Wu, Amy Greenwald, and Ronald Parr for pointing out a mistaken condition in Proposition 3.1, which has been corrected in the current ArXiv version.
We thankfully acknowledge funding from the Canada CIFAR AI Chairs program, the Reinforcement Learning and Artificial Intelligence (RLAI) laboratory, the Alberta Machine Intelligence Institute (Amii), and the Natural Sciences and Engineering Research Council (NSERC) of Canada. 

\bibliography{example_paper}
\bibliographystyle{icml2024}

\newpage
\appendix
\onecolumn
\section{Appendix}
\subsection{Over-parameterized Target TD Convergence Proof} \label{A:convergence}
In this proposition, we study a sufficient conditon for the convergence of over-paramterized TD (OTD).
\begin{proposition}
    A sufficient condition for the matrix $I - \eta (M-\gamma N) M^{\top} D_k $ to have a spectral radius of less than one would be to have orthonormal feature vectors and states showing up uniformly.
\end{proposition}

\begin{proof}
    Thanks to orthornormality, $MM^{\top} = I$ is the identity matrix and $NM^{\dagger} = H \hat{P}H^{\top}$.

    Thus, the spectral radius equals
    \begin{align*}
        \rho(I - \eta (M-\gamma N) M^{\top} D_k) &= \rho(I - \eta D_k (I - \gamma H \hat{P}H^{\top}) )\\
        &= \lVert I - \eta D_k +\eta D_k \gamma H \hat{P}H^{\top} \rVert_{\infty}\\
        &\le \lVert I - \eta D_k\rVert_{\infty} + \lVert \eta D_k\rVert_{\infty} \lVert\gamma H \hat{P}H^{\top} \rVert_{\infty} \\
        & \le 1- \eta \min_{s,a} \hat{\lambda}(s,a) + \gamma \eta \max_{s,a} \hat{\lambda}(s,a)\\
        & = 1- \eta \frac{1}{k} + \gamma \eta \frac{1}{k}<1.
    \end{align*}
\end{proof}

Next, we start the proof for the main result, the convergence of OTTD.
\begin{lemma}
    When $M$ is of full rank, $W=NM^{\dagger}$ has some sub-multiplicative norm smaller than or equal to one and the learning rate satisfies $\rho(\eta MM^{\top}D_k)<1$, there exists an integer $\Bar{m}$ such that for all target parameter update step $m \ge \Bar{m}$, the spectral radius of $\gamma W + (I - \gamma W)(I - \eta M M^{\top} D_k)^m$ is strictly smaller than one.
    \label{thm:radius_less_one}
\end{lemma}

\begin{proof}
    Let $U\Lambda U^{-1}$ be the eigen-decomposition of $MM^{\top} D_k$ where $\Lambda = diag(\lambda_1, \cdots , \lambda_k)$ with $\frac{1}{\eta}>\lambda_1 \ge \cdots \lambda_k>0$. All eigenvalues are positive due to the symmetry of $D_k^{\frac{1}{2}}MM^{\top}D_k^{\frac{1}{2}}$ and the full rank of $M$. Denote the matrix $(I- \eta M M^{\top} D_k)^m$ as $A$. Thus, 
    \begin{equation}
        A = (I- \eta M M^{\top} D_k)^m = (I - \eta U \Lambda U^{-1})^m = (U U^{-1} - \eta U \Lambda U^{-1})^m = U (I -\eta \Lambda)^mU^{-1}.
    \end{equation}

    The diagonal matrix $(I -\eta \Lambda)^m$ converges to zero as $m$ tends to infinity. Consequently, any norm of the matrix goes to zero as well. Such that there exists a constant $\Bar{m}$, for all step $m \ge \Bar{m}$, the norm of $(I -\eta \Lambda)^m$ is bounded by $\frac{1-\gamma}{1+\gamma} \frac{1}{\lVert U \rVert \lVert U^{-1} \rVert}$. Therefore, the norm of $A$ is bounded as
    \begin{align*}
        \lVert A \rVert &\le \lVert U \rVert   \lVert (I -\eta \Lambda)^m \rVert  \lVert U^{-1} \rVert\\
        & < \frac{1-\gamma}{1+\gamma}.
    \end{align*}

    With the upper bound of the matrix norm, we can proceed and show the desired matrix has spectral radius $\rho(\gamma W + (I - \gamma W)(I - \eta M M^{\top} D_k)^m)$ less than one for all $m \ge \Bar{m}$.
    \begin{align*}
        \rho(\gamma W + (I - \gamma W)(I - \eta M M^{\top} D_k)^m) &\le \lVert \gamma W + (I - \gamma W)(I - \eta M M^{\top} D_k)^m \rVert \\
        &\le \lVert \gamma W\rVert  + \lVert(I - \gamma W)\rVert \lVert(I - \eta M M^{\top} D_k)^m \rVert\\
        & \le \gamma + (1+\gamma)\lVert(I - \eta M M^{\top} D_k)^m \rVert <1.
    \end{align*}
    
\end{proof}

\begin{proposition}
    Each $m$ steps of over-parameterized target TD updates can be combined to
    \begin{equation}
        \theta_{(n+1)m} = (I-\eta M^{\top} BD_k (M-\gamma N ))\theta_{nm} + \eta M^{\top}BD_kR, 
    \end{equation}
    where $B = \sum_{i=0}^{m-1} (I - \eta D_kMM^{\top})^i$.
    \label{thm:comb_update_rule}
\end{proposition}
\begin{proof}
The original update rule can be written $m$ times recursively and gives that
    \begin{align*}
    \theta_{(n+1)m} &=(I - \eta M^{\top}D_kM ) \theta_{(n+1)m-1}  + \eta M^{\top}D_k(R + \gamma  N  \theta_{\mathrm{targ},nm}) \nonumber \\
    &= (I - \eta M^{\top}D_kM )^m \theta_{nm} + \sum_{i=0}^{m-1} (I - \eta M^{\top}D_kM )^i\eta M^{\top}D_k(R + \gamma N  \theta_{\mathrm{targ},nm})\nonumber\\
    &= (I - \eta M^{\top}D_kM )^m \theta_{nm} + \sum_{i=0}^{m-1} (I - \eta M^{\top}D_kM )^i\eta M^{\top}D_k(R + \gamma N  \theta_{nm} ). 
\end{align*}

The update rule includes powers of $I - \eta M^{\top}D_kM$ can be reduced to terms dependent on matrix $B=\sum_{i=0}^{m-1} (I - \eta D_kMM^{\top})^i$.

The $m$-th power of the matrix can be reduced as
\begin{align}
    (I - \eta M^{\top} D_k M)^m &= (I - \eta M^{\top} D_k M)^{m-1} (I - \eta M^{\top} D_k M)\nonumber \\
    &= (I - \eta M^{\top} D_k M)^{m-1} - (I - \eta M^{\top} D_k M)^{m-1}\eta M^{\top} D_k M\nonumber \\
    &= (I - \eta M^{\top} D_k M)^{m-1} - \eta M^{\top} (I - \eta D_k M M^{\top})^{m-1} D_k M\nonumber \\
    &= \cdots \nonumber \\
    &= I - \eta M^{\top} \sum_{i=0}^{m-1} (I - \eta D_k M M^{\top})^i D_k M\nonumber \\
    &= I - \eta M^{\top}BD_kM. \label{eq:student_part_I}
\end{align}

Also, notice that sum of matrix power times $M^{\top}$ can be simplied as
\begin{align*}
    \sum_{i=0}^{m-1} (I - \eta M^{\top}D_kM )^i\eta M^{\top} & = \sum_{i=0}^{m-1} (\eta M^{\top} - \eta M^{\top}D_kM \eta M^{\top} )^i \\
    &= \eta M^{\top} B.
\end{align*}

Thus, we have
\begin{align*}
    \theta_{(n+1)m} 
    &= (I - \eta M^{\top}BD_kM) \theta_{nm} + \sum_{i=0}^{m-1} (I - \eta M^{\top}D_kM )^i\eta M^{\top}D_k(R + \gamma N  \theta_{nm} )\nonumber\\
    &=(I - \eta M^{\top}BD_kM)\theta_{nm} + \eta M^{\top} B D_k (R + \gamma N \theta_{nm})\nonumber\\
    &=(I - \eta M^{\top}BD_k(M -\gamma N))\theta_{nm} +\eta M^{\top} BD_k R. 
\end{align*}

\end{proof}
\clearpage

\begin{theorem}
    When $M$ has full rank, $W=NM^{\dagger}$ has some sub-multiplicative norm smaller than or equal to one and the learning rate satisfies $\rho(\eta MM^{\top}D_k)<1$, there exists an integer $\bar{m}$ such that for all update steps of the target parameter $m \ge \bar{m}$, the parameter of over-parameterized target TD always converge to
    \[\theta_{target}^* = M^{\dagger}(I-\gamma W)^{-1}R + (I - M^{\dagger}M+ M^\dagger (I-\gamma W)^{-1} \gamma N (I - M^{\dagger}M))\theta_0,\]
    where $\theta_0$ is the initial point.
    \label{thm:full_OTTD_initial_point}
\end{theorem}

\begin{proof}
    A simple recursive argument shows that for any $\theta \in \mathbb{R}^d$, the update rule given in Proposition \ref{thm:comb_update_rule}.
\begin{align}
    &\theta_{(n+1)m} = (I-\eta M^{\top} BD_k (M-\gamma N ))\theta_{nm} + \eta M^{\top}BD_kR \nonumber\\
    &= (I-\eta M^{\top} BD_k (M-\gamma N ))^{n+1}\theta_{0}
    + \sum_{i=0}^n(I-\eta M^{\top} BD_k (M-\gamma N ))^{i} \eta M^{\top}BD_kR\nonumber \\
    &= \underbrace{(I-\eta M^{\top} BD_k (M-\gamma N ))^{n+1}\theta_{0}}_\text{\clap{term 1}}
    + \underbrace{\eta M^{\top}B D_k \sum_{i=0}^n(I- (I-\gamma W) \eta M M^{\top}BD_k)^{i} R}_\text{\clap{term 2}}.
    \label{eq:update_rule_with_intial}
\end{align}
where $B = \sum_{i=0}^{m-1} (I - \eta D_kMM^{\top})^i$. Also, the last line uses that $(M-\gamma N) M^\top = (I - \gamma N M^\dagger) M M^\top$, where we further denote $NM^\dagger$ to $W$.

We work on the second term independent of the initial point first and simplify the summation of matrix powers. Reversing the recursive step in Equation \ref{eq:student_part_I}, $\eta M M^{\top}BD_k$ can be expressed as $I - (I-\eta M M^{\top} D_k)^m$. Thus, the matrix in the summation can be rewritten as
\begin{align*}
    I- (I-\gamma W) \eta M M^{\top}BD_k =  \gamma W +(I-\gamma W)(I-\eta M M^{\top} D_k)^m
\end{align*}

By Lemma \ref{thm:radius_less_one}, there exists $\Bar{m}$ such that for all $m \ge \Bar{m}$, $I- (I-\gamma W) \eta M M^{\top}BD_k $ has spectral radius less than one. Thus, by properties of Neumann series, the summation of its matrix powers converges, that is
\begin{equation}
    \sum_{i=0}^n(I- (I-\gamma W) \eta M M^{\top}BD_k)^{i} \to [(I-\gamma W) \eta M M^{\top}BD_k]^{-1}, \quad \text{ as } n \to \infty.
\end{equation}

Notice that $B$ is invertible, since $(I- \eta D_k MM^{\top})^{\top}$ is strictly diagonal dominant and thus has positive eigenvalues. Then its powers and sum of powers also have positive eigenvalues and thus are invertible. Also, $MM^{\top}$ is invertible, thanks to the full rank of $M$. Therefore,
\begin{align}
    & \lim_{n \to \infty} \sum_{i=0}^n \eta M^{\top}B D_k (I- (I-\gamma W) \eta M M^{\top}BD_k)^{i} R\nonumber \\
    & = \eta M^{\top}B D_k (\eta M M^{\top}BD_k)^{-1} (I-\gamma W)^{-1} R\nonumber \\
    &= M^{\dagger} (I-\gamma W)^{-1} R.
\end{align}

Now we start dealing with the first term in Equation \ref{eq:update_rule_with_intial} and study how the fixed point depends on the initial value.

\begin{align*}
    &(I-\eta M^{\top} B D_k (M-\gamma N ))^{n+1} \nonumber \\
    &= I - \sum_{i=0}^n (I-\eta M^{\top} B D_k (M-\gamma N ))^{i} \eta M^{\top} B D_k (M-\gamma N ) \nonumber \\
    &= I - \sum_{i=0}^n \eta M^{\top} B D_k (I- (I-\gamma W) \eta M M^{\top} B D_k)^i(M-\gamma N ) \nonumber \\
    & \xrightarrow{n \to \infty} I -  M^{\top} B D_k ( M M^{\top} B D_k)^{-1}(I-\gamma W)^{-1} (M-\gamma N M^\dagger M - \gamma N (I - M^\dagger M) ) \nonumber \\
    &= I - M^{\dagger}M+M^\dagger (I-\gamma W)^{-1} \gamma N (I - M^\dagger M).
\end{align*}
The second line uses the same recursive trick in Equation \ref{eq:student_part_I}.

In conclusion, the parameter $\theta_{nm}$ converges to $ M^{\dagger} (I-\gamma W)^{-1} R$, as update step $n$ goes to infinity, when the initial point equals zero.
\end{proof}

Next, we prove the proposition for the special case with expected updates. The data distribution used for expected updates covers the entire state-action space.

\begin{proposition} 
For the over-parameterized regime $d>k$, there always exists a learning rate $\eta$ and an integer $\bar{m}$ such that for all update window sizes of the target parameters $m \ge \bar{m}$, the parameter of OTTD converges to 
    \[\theta_{\mathrm{TD}}^* = \Phi^{\dagger}(I-\gamma P_{\pi})^{-1}R + (I - \Phi^{\dagger}\Phi)\theta_0,\]
    where $\theta_0$ is the initial parameter of OTTD.
\end{proposition}

\begin{proof}
These updates assume ideal offline data encompassing all state-action pairs such that $k=|\cS||\cA|$, $M=\Phi$, $N=P_\pi \Phi$ and $D= \text{diag}(\lambda)$. 
It can be easily verified that for the expected update, we have $NM^\dagger = P_\pi \Phi \Phi^\dagger=P_\pi$, which naturally satisfies that it has a sub-multiplicative norm smaller than or equal to one with each row summing up to one.
Hence, as demonstrated in \cref{thm:target_TD_convergence}, OTTD converges with a proven guarantee.
\end{proof}

\subsection{Bound of Value Estimation Error}\label{A:error}
\begin{theorem}
    Given a dataset $\mathcal{D}$ and the optimal fixed point $\theta_{TD}^* = M^{\dagger}(I-\gamma N M^{\dagger})^{-1} R$, when $\lVert N M^{\dagger} \rVert_{\infty}<1$, with probability at least $1-\delta$, for $\theta^* = argmin_{\theta} \lVert \Phi \theta -q \rVert_{\infty}$,
    \begin{align}
       &\lVert \Phi \theta_{TD}^*-q \rVert_{\infty} \\
       &\le \frac{\lVert \Phi M^{\dagger} \rVert_{\infty}  }{(1-\gamma)^2 \sqrt{\min_{s,a}n(s,a)}}\sqrt{\log(\frac{2k |\mathcal{A}|}{\delta})} + \frac{\lVert \Phi M^{\dagger} \rVert_{\infty}}{1-\gamma} \lVert \Phi(I- M^{\dagger}M) \theta^* \rVert_{\infty}+\frac{2 \lVert \Phi M^{\dagger} \rVert_{\infty}}{1-\gamma} \lVert \Phi \theta^* -q \rVert_{\infty}.
    \end{align}
\end{theorem}

\begin{proof}
    \begin{align}
        \lVert \Phi \theta_{TD}^*-q \rVert_{\infty} &= \lVert \Phi M^{\dagger}M\theta_{TD}^*-q \rVert_{\infty} \nonumber\\
        &\le \underbrace{\lVert \Phi M^{\dagger}M\theta_{TD}^*- \Phi M^{\dagger} Hq\rVert_{\infty}}_\text{\clap{term 1}} + \underbrace{\lVert \Phi M^{\dagger} Hq - q\rVert_{\infty}}_\text{\clap{term 2}}. \label{eq:gen_err_first_decomp}
    \end{align}

    Define $\theta^* = argmin_{\theta} \lVert \Phi \theta -q \rVert_{\infty}$. By adding intermediate term, we can bound the second term
    \begin{align}
        &\lVert \Phi M^{\dagger} Hq - q\rVert_{\infty} \label{eq:intermediate_step}\\
        &=  \lVert \Phi M^{\dagger} Hq - \Phi M^{\dagger} M \theta^* +  \Phi\theta^* -q - \Phi (I - M^{\dagger} M)\theta^* \rVert_{\infty} \nonumber\\
        &\le  \lVert \Phi M^{\dagger} H (q - \Phi \theta^*)\rVert_{\infty} +
        \lVert  \Phi\theta^* -q \rVert_{\infty} + \lVert  \Phi (I - M^{\dagger} M)\theta^* \rVert_{\infty} \nonumber\\
        &\le (\lVert \Phi M^{\dagger} \rVert_{\infty}+1) \lVert \Phi \theta^* -q \rVert_{\infty} + \lVert \Phi(I- M^{\dagger}M) \theta^* \rVert_{\infty}. \nonumber
    \end{align}
    
    \begin{enumerate}
    \item $A =\lVert H \hat{P} (\Phi M^{\dagger}Hq-q) \rVert_{\infty} \le 2 \lVert \Phi \theta^*-q \rVert_{\infty}+\lVert \Phi (I- M^{\dagger}M)^*\rVert_{\infty}$.
    \item Define $\epsilon_{est} = \lVert \gamma H (P-\hat{P} )q \rVert_{\infty}+\lVert R - Hr \rVert_{\infty}$.
    \item $B = \lVert H \hat{P} \Phi M^{\dagger}(M\theta_{TD}^*-Hq) \rVert_{\infty} \le \frac{1}{1-\gamma}\epsilon_{est} + \frac{\gamma}{1-\gamma}A$.
\end{enumerate}
    The first equation repeats the steps to bound the term in Equation \ref{eq:intermediate_step} and uses $\lVert H \hat{P} \Phi M^{\dagger}\rVert_{\infty}= \lVert N M^{\dagger}\rVert_{\infty} \le1$.
    
    The third equation further uses the extension of $M \theta^*$ and $Hq$ following the Bellman update:
    \begin{align*}
        M \theta_{TD}^* &= R+ \gamma W M \theta_{TD}^* = R + \gamma H \hat{P} \Phi M^{\dagger} M \theta_{TD}^*,\\
        Hq &= Hr + \gamma HPq = Hr + \gamma H \hat{P} \Phi M^{\dagger}Hq +\gamma H \hat{P}(q- \Phi M^{\dagger}Hq) + \gamma H (P-\hat{P})q.
    \end{align*}

    Next, we can bound the first term in Equation \ref{eq:gen_err_first_decomp} by extending $M \theta_{TD}^*$ and $Hq$.
    \begin{align}
        &\lVert\Phi M^{\dagger}M \theta_{TD}^*-\Phi M^{\dagger}Hq\rVert_{\infty} \nonumber \\
        & \le \lVert\Phi M^{\dagger}( R-Hr +\gamma H (P-\hat{P})q) + \Phi M^{\dagger} \gamma H \hat{P} \Phi M^{\dagger} (M \theta_{TD}^*-Hq )+\Phi M^{\dagger}\gamma H \hat{P}(q- \Phi M^{\dagger}Hq)\rVert_{\infty}\nonumber\\
        &\le \lVert \Phi M^{\dagger} \rVert_{\infty}\epsilon_{est}+\gamma\lVert \Phi M^{\dagger} \rVert_{\infty}A+\gamma\lVert \Phi M^{\dagger} \rVert_{\infty}B\nonumber\\
         & \le \frac{\lVert \Phi M^{\dagger} \rVert_{\infty}}{1-\gamma}\epsilon_{est}+\frac{\gamma \lVert \Phi M^{\dagger} \rVert_{\infty}}{1-\gamma}A \nonumber \\
         &\le \frac{\lVert \Phi M^{\dagger} \rVert_{\infty}}{1-\gamma}\epsilon_{est}+\frac{2\gamma \lVert \Phi M^{\dagger} \rVert_{\infty}}{1-\gamma} \lVert \Phi \theta^*-q \rVert_{\infty}+\frac{\gamma \lVert \Phi M^{\dagger} \rVert_{\infty}}{1-\gamma}\lVert \Phi (I- M^{\dagger}M)^*\rVert_{\infty}.
    \end{align}

    Thus, combining the bounds on the first and the second term in Equation \ref{eq:gen_err_first_decomp}, we gain
    \begin{align}
        &\lVert \Phi \theta_{TD}^*-q \rVert_{\infty} \nonumber\\
        &\le   \frac{\lVert \Phi M^{\dagger} \rVert_{\infty}}{1-\gamma}\epsilon_{est}+(\frac{\gamma \lVert \Phi M^{\dagger} \rVert_{\infty}}{1-\gamma}+1)\lVert \Phi (I- M^{\dagger}M)^*\rVert_{\infty} + (\lVert \Phi M^{\dagger} \rVert_{\infty}+1+\frac{2\gamma \lVert \Phi M^{\dagger} \rVert_{\infty}}{1-\gamma}) \lVert \Phi \theta^* -q \rVert_{\infty} \nonumber\\
        & \le \frac{\lVert \Phi M^{\dagger} \rVert_{\infty}}{1-\gamma} \epsilon_{est} +\frac{ \lVert \Phi M^{\dagger} \rVert_{\infty}}{1-\gamma} \lVert \Phi(I- M^{\dagger}M) \theta^* \rVert_{\infty} + \frac{2 \lVert \Phi M^{\dagger} \rVert_{\infty}}{1-\gamma} \lVert \Phi \theta^* -q \rVert_{\infty}, \label{eq:bound_with_stat_err}
    \end{align}
    since $\lVert \Phi M^{\dagger} \rVert_{\infty} \ge 1$.

    $\epsilon_{est}$ is bounded by Hoeffding's inequality and a union bound. We have with probability at least $1-\delta$, for any state $s$ and action $a$ with $d_k(s)>0$ in the dataset, 
    \begin{align*}
        | (P_{s,a} - \hat{P}_{s,a})^{\top} q | &\le \frac{1}{(1-\gamma) \sqrt{2 n(s,a)}} \sqrt{log(\frac{2k|\mathcal{A}|}{\delta})}.\\
        |R(s,a) - r(s,a)| &\le \frac{1}{\sqrt{2 n(s,a)}} \sqrt{log(\frac{2k|\mathcal{A}|}{\delta})}.
    \end{align*}
    
    Combining all terms, we gain the bound.

    As the dataset gradually covers the entire state space and gain the actual transition matrix and reward, that is, $\hat{P}_{\mu} \to P_{\mu}$, $R \to r$ and $M \to \Phi$, $\epsilon_{est} \to 0$ and $\Phi M^{\dagger} \to I$. Only $\frac{2}{1-\gamma} \lVert \Phi \theta^* -q \rVert_{\infty}$ is left in the bound.
\end{proof}

\subsection{Nomalized Importance Sampling Results}\label{A:NIS}

\begin{proposition}
    The transition probability estimator $\hat{P}_{\pi}(s',a'|s,a)$ is consistent and tends to $P_{\pi}(s',a'|s,a)$ almost surely as $n(s,a) \to \infty$, which is the counts of the current state-action pair $(s,a)$.
\end{proposition}

\begin{proof}
    Each time a current state-action pair $(s,a)$ shows up, we can define two random variables:
    \begin{align}
        X &= \rho(a'|s')\mathds{1}[S=s,A=a,S'=s',A'=a'],\\
        Y &= \sum_{\Tilde{s},\Tilde{a}}\rho(\Tilde{a}|\Tilde{s})\mathds{1}[S=s,A=a,S'=\Tilde{s},A'=\Tilde{a}].
    \end{align}

    These two random variables are sampled $n(s,a)$ times, labeled from $j=1$ to $n(s,a)$. Since each next action can be sampled from a different behaviour policy, $X_1,\cdots, X_{n(s,a)}$ are not sampled from the same distribution but are independent conditioned on the current state-action pair, similar for $Y_j$, $j=1,\cdots,n(s,a)$.

    Our estimator $\hat{P}_{\pi}(s',a'|s,a)$ can be seen as a ratio of the average of these random variables, that is, 
    \begin{equation}
        \hat{P}_{\pi}(s',a'|s,a) = \cfrac{\sum_{j=1}^{n(s,a)}X_j}{\sum_{j=1}^{n(s,a)}Y_j}.
    \end{equation}

    Notice that for $X_j$, the mean equals
    \begin{align}
        \mathbb{E} X_j &= \rho(a'|s')\mathbb{E} \left [\mathds{1}[S=s,A=a,S'=s',A'=a'] \right ] \\
        &= \cfrac{\pi(a'|s')}{\mu_i(a'|s')}\mathbb{P}(S=s,A=a,S'=s',A'=a') \\
        &= P_{\pi}(s',a'|s,a).
    \end{align}
    Similarly, we gain that $\mathbb{E} Y_j=1$.

    Denote $M_{\rho} := \max_{i} \rho(A_i'|S_i')$. The variance of $X_j$ and $Y_j$ are bounded by $M_{\rho}^2$. 
    
    Then by strong large law of number for martingales, $\frac{1}{n(s,a)}\sum_{j=1}^{n(s,a)}X_j \to P_{\pi}(s',a'|s,a)  $ and $ \frac{1}{n(s,a)} \sum_{j=1}^{n(s,a)}Y_j\to 1$ almost surely. Since $\mathbb{P}(Y_j>0) =1$, their ratio $\hat{P}_{\pi}(s',a'|s,a) \to P_{\pi}(s',a'|s,a) $ a.s..
\end{proof}

\begin{theorem}
    Given a batch of trajectories under some behaviour policies, if $M$ has full rank and the learning rate satisfies that $\eta < \frac{1}{\rho(MM^{\top}D_k)}$, there exists an integer $\bar{m}$ such that for all update steps of the target parameter $m \ge \bar{m}$, the parameter of corrected over-parameterized target TD always converge to
    \[\theta_{TD,\mathrm{NIS}}^* = M^{\dagger}(I-\gamma N_{\mathrm{NIS}M^{\dagger}} )^{-1}R + (I - M^{\dagger}M)\theta_0,\]
    where $\theta_0$ is the initial point.
\end{theorem}

\begin{proof}
    The update rule equals $\theta_{t+1} = \theta_t - \eta M^{\top}D_k \left [M\theta_t - (R + \gamma  N_{\mathrm{NIS}} \theta_{\mathrm{targ},m \lfloor \frac{t}{m} \rfloor}) \right]$, which is of the same form as Theorem \ref{thm:target_TD_convergence}.

    As long as the assumptions for Theorem \ref{thm:target_TD_convergence} are satisfied, the convergence is proved. The assumption requires the projected coefficient matrix $N_{\mathrm{NIS}} M^{\dagger}  = H\hat{P}_{\pi,\mathrm{NIS}}\Phi M^{\dagger}$ to satisfy the condition of having a norm less than one. Here, $W_{\mathrm{NIS}} $ equals the non-zero square matrix in $\hat{P}_{\pi}$ and is also stochastic. Therefore, the its infinity norm is less than one and the condition is satisfied.
\end{proof}

\subsection{Value Estimation Error Bound with NIS for Episodic Tasks}\label{A:NIS_error}
First, we state a lemma from Sharoff and colleagues (2020) \nocitenoh{sharoff2020farewell} for the concentration inequality of average ratios.
\begin{lemma}
    Let $X$, $Y$ be possibly dependent random variables with joint distribution $P$. Consider a sample $(X_1, Y_1), \ldots, (X_n, Y_n)$ of independent copies of $(X, Y) \sim P$. Assume that $X$ takes values in $[0, 1]$ and $Y$ takes values in $[1, B]$. Define $\mu_Y := E[Y]$ and let $\bar{X}$ denote the sample mean of $X_1, \ldots, X_n$ (likewise for $\bar{Y}$ and $Y_1, \ldots, Y_n$). For any choice of $\delta \in [0, 1]$, we have with probability at least $1 - \delta$ over the sample,
\[
\left |\frac{\bar{X}}{\bar{Y}} - \frac{E[X]}{E[Y]} \right |\leq \sqrt{\frac{(B-1) \log \frac{4}{\delta}}{2n}} + \frac{2(B-1) \log \frac{4}{\delta}}{3\mu_Y n} + \sqrt{\frac{ \log \frac{4}{\delta}}{2n}} \le \frac{ \max \{1,B-1\}\log \frac{4}{\delta}}{\min\{\mu_Y,1\} \sqrt{2n}}.
\]
\end{lemma}

\begin{corollary}
    Given a dataset $\mathcal{D}$ of episodic trajectory data collected under a behaviour policy $\mu$ and the optimal fixed point $\theta_{TD}^* = M^{\dagger}(I-\gamma H \hat{P}_{\pi}\Phi M^{\dagger})^{-1} R$, when $\lVert H \hat{P}_{\pi}\Phi M^{\dagger} \rVert_{\infty}<1$ and $\mu$ covers the support of the target policy $\pi$, with probability at least $1-\delta$,
    \begin{align}
        &\lVert \Phi \theta_{TD}^*-q_{\pi} \rVert_{\infty} \nonumber \\
        &\le \epsilon_{stat}'+\epsilon_{projection}+\epsilon_{approx}.
    \end{align}
    Denote $\rho_M:=\cfrac{\max_{(s',a') \in \mathcal{D}} \rho(a'|s')}{\min_{(s',a') \in \mathcal{D}} \rho(a'|s')} $.
    
    $\epsilon_{stat}' = \frac{ \lVert \Phi M^{\dagger} \rVert_{\infty}   \rho_M}{(1-\gamma)^2 \sqrt{\min_{(s,a)}n(s,a)}}\log(\frac{4k|\mathcal{A}|}{\delta})\max\{\rho_M-1,1\}$. 
    
    $\epsilon_{projection}$ and $\epsilon_{approx}$ are the same as in Theorem \ref{thm:target_td_gen_error}.
\end{corollary}
\begin{proof}
As shown in Equation \ref{eq:bound_with_stat_err},
\begin{align}
    &\lVert \Phi \theta_{TD}^*-q \rVert_{\infty} \nonumber\\
    & \le \frac{\lVert \Phi M^{\dagger} \rVert_{\infty}}{1-\gamma} \epsilon_{est} +\frac{ \lVert \Phi M^{\dagger} \rVert_{\infty}}{1-\gamma} \lVert \Phi(I- M^{\dagger}M) \theta^* \rVert_{\infty} + \frac{2 \lVert \Phi M^{\dagger} \rVert_{\infty}}{1-\gamma} \lVert \Phi \theta^* -q \rVert_{\infty}.
\end{align}

Compared to the fixed point of OTTD with sampled target actions, only the transition probability estimation $\hat{P}$ is changed. This only influences the statistical error, $\epsilon_{est} = \lVert \gamma H (P-\hat{P} )q \rVert_{\infty}+\lVert R - Hr \rVert_{\infty}$, which is bounded by the above lemma and a union bound. 

\begin{equation}
    \hat{P}_{\pi,\mathrm{NIS}}(s,a,s',a') =  \frac{\sum_{i=0}^n \rho(a'|s')\mathds{1}[S_i=s,A_i=a,S_i'=s',A_i'=a']}{\sum_{i=1}^n \sum_{\Tilde{s},\Tilde{a}}\rho(\Tilde{a}|\Tilde{s})\mathds{1}[S_i=s,A_i=a,S_i'=\Tilde{s},A_i'=\Tilde{a}]}.
\end{equation}
As defined in \ref{A:NIS},
\begin{align}
    X &= \rho(a'|s')\mathds{1}[S=s,A=a,S'=s',A'=a'],\\
    Y &= \sum_{\Tilde{s},\Tilde{a}}\rho(\Tilde{a}|\Tilde{s})\mathds{1}[S=s,A=a,S'=\Tilde{s},A'=\Tilde{a}].
\end{align}

$\frac{X}{\max_{s',a'} \rho(a'|s')} \in [0,1]$ and $\frac{Y}{\min_{s',a'} \rho(a'|s')} \in [1,\rho_M]$.

We have with probability at least $1-\delta$, for any state $s$ and action $a$ with $d_k(s)>0$ in the dataset, 
\begin{align*}
    | (P_{s,a} - \hat{P}_{s,a})^{\top} q | &\le \frac{  \rho_M \max\{\rho_M-1,1\}}{ (1-\gamma) \sqrt{2n(s,a)}} log(\frac{4k|\mathcal{A}|}{\delta}).\\
    |R(s,a) - r(s,a)| &\le \frac{1}{\sqrt{2 n(s,a)}} \sqrt{log(\frac{2k|\mathcal{A}|}{\delta})}.
\end{align*}    
\end{proof}

\subsection{Continuing Tasks' Error Bound}\label{A:cont_error}

\begin{corollary}
    Given a dataset with trajectory data under a behaviour policy $\mu$ and the optimal fixed point $\theta_{TD}^* = M^{\dagger}(I-\gamma H \hat{P}_{\pi}\Phi M^{\dagger})^{-1} R$, when $\lVert \hat{P}_{\pi}\Phi M^{\dagger} \rVert_{D_{\pi}}<1$ and $\mu$ covers the support of the target policy $\pi$, with probability at least $1-\delta$, under the stationary distribution $d_{\pi}$,
    \begin{align}
       &\lVert \Phi \theta_{TD}^*-q_{\pi} \rVert_{D_{\pi}} \\
        &\le \epsilon_{stat}''+\epsilon_{projection}''+\epsilon_{approx}''.
    \end{align}
    Denote $\rho_M:=\cfrac{\max_{(s',a') \in \mathcal{D}} \rho(a'|s')}{\min_{(s',a') \in \mathcal{D}} \rho(a'|s')} $, and $C:= \lVert \Phi M^{\dagger} H \rVert_{D_{\pi}}$.
    
    $\epsilon_{stat}'' = \frac{ C   \rho_M}{(1-\gamma)^2 \sqrt{\min_{(s,a)}n(s,a)}}\log(\frac{4k|\mathcal{A}|}{\delta})\max\{\rho_M-1,1\}+ \frac{C  }{(1-\gamma)^2 }$. 
    
    Let $\theta^*$ be defined as $\theta^* = argmin_{\theta} \lVert \Phi \theta -q_{\pi} \rVert_{D_{\pi}}$. Then, $\epsilon_{projection}'' = \frac{C(1+\gamma \lVert \hat{P}_{\pi,\mathrm{NIS}}\rVert_{D_{\pi}})}{1-\gamma} \lVert \Phi(I- M^{\dagger}M) \theta^* \rVert_{D_{\pi}}$, and 
    
    $\epsilon_{approx}'' = \frac{C(2+\gamma \lVert \hat{P}_{\pi,\mathrm{NIS}}\rVert_{D_{\pi}})}{1-\gamma} \lVert \Phi \theta^* -q_{\pi} \rVert_{D_{\pi}}$. 
    \label{thm:infinite_gen_error}
\end{corollary}

\begin{proof}
The proof is similar as the one in \ref{A:error} except for a norm change.
    \begin{align}
        \lVert \Phi \theta_{TD}^*-q \rVert_{D_{\pi}} &= \lVert \Phi M^{\dagger}M\theta_{TD}^*-q \rVert_{D_{\pi}} \nonumber\\
        &\le \underbrace{\lVert \Phi M^{\dagger}M\theta_{TD}^*- \Phi M^{\dagger} Hq\rVert_{D_{\pi}}}_\text{\clap{term 1}} + \underbrace{\lVert \Phi M^{\dagger} Hq - q\rVert_{D_{\pi}}}_\text{\clap{term 2}}. 
    \end{align}

    Define $\theta^* = argmin_{\theta} \lVert \Phi \theta -q \rVert_{D_{\pi}}$. By adding intermediate term, we can bound the second term
    \begin{align*}
        &\lVert \Phi M^{\dagger} Hq - q\rVert_{D_{\pi}} \\
        &\le  \lVert \Phi M^{\dagger} H (q - \Phi \theta^*)\rVert_{D_{\pi}} +
        \lVert  \Phi\theta^* -q \rVert_{D_{\pi}} + \lVert  \Phi (I - M^{\dagger} M)\theta^* \rVert_{D_{\pi}}\\
        &\le (\lVert \Phi M^{\dagger}H \rVert_{D_{\pi}}+1) \lVert \Phi \theta^* -q \rVert_{D_{\pi}} + \lVert \Phi(I- M^{\dagger}M) \theta^* \rVert_{D_{\pi}}.
    \end{align*}
    
    \begin{enumerate}
    \item $A =\lVert \hat{P} (\Phi M^{\dagger}Hq-q) \rVert_{D_{\pi}} \le (1+\lVert \hat{P}_{\pi,\mathrm{NIS}}\rVert_{D_{\pi}}) \lVert \Phi \theta^*-q \rVert_{D_{\pi}}+\lVert \hat{P}_{\pi,\mathrm{NIS}}\rVert_{D_{\pi}} \lVert \Phi (I- M^{\dagger}M)^*\rVert_{D_{\pi}}$.
    \item Define $\epsilon_{est} = \lVert \gamma (P-\hat{P}_{\pi,\mathrm{NIS}} )q \rVert_{D_{\pi}}+\lVert R - r \rVert_{D_{\pi}}$.
    \item $B = \lVert \hat{P}_{\pi,\mathrm{NIS}} \Phi M^{\dagger}(M\theta_{TD}^*-Hq) \rVert_{D_{\pi}} \le \frac{1}{1-\gamma}\epsilon_{est} + \frac{\gamma}{1-\gamma}A$.
\end{enumerate}
   These three statements are almost the same as in the proof in \ref{A:error} except the norm difference.

    When bounding $\epsilon_{est}$, we need to count the error from loop transitions. It adds in wrong transitions for the last state-action pair $(s_T,a_T)$ and thus, we can only bound as
    \begin{equation}
        | (P_{s_T,a_T} - \hat{P}_{s_T,a_T})^{\top} q | \le \frac{2 }{(1-\gamma)}.
    \end{equation}
    In the norm, this term is weighted by the stationary distribution $d_{\pi}(s_T,a_T)$.
    For state-action pair without additional loop transitions, we use the lemma from Sharoff and colleagues (2020) stated in \ref{A:NIS_error} and a union bound.
    
    Combining all terms, we gain the bound.
\end{proof}

\subsection{Proof of  Q-learning Convergence}\label{A:Q}
\begin{theorem}
    When $ \lVert \Phi'^{\top} M^{\dagger} \rVert_{\infty}<\frac{1}{\gamma}$, there exists an integer $\bar{m}$ such that for all update steps of the target parameter $m \ge \bar{m}$, parameter of over-parameterized target Q-learning converges to a fixed point $\theta^* = M^{\dagger}\hat{q}^*+(I - M^{\dagger}M) \theta_0$, where $\hat{q}^* \in \mathbb{R}^k$ satisfies
    \[\hat{q}^* = R + \gamma  \hat{P} \begin{pmatrix}
\lVert \Phi_1 M^{\dagger} \hat{q}^* \rVert_{\infty} \\
\lVert \Phi_2 M^{\dagger} \hat{q}^* \rVert_{\infty} \\
\cdots \\
\lVert \Phi_{\mathcal{S}} M^{\dagger} \hat{q}^*\rVert_{\infty} \\
\end{pmatrix},\]
    and $\theta_0$ is the initial point.
\end{theorem}

\begin{proof}
We proceed in the same way to Theorem 1.1. The update rule for student parameter every $m$ steps equal
\begin{equation}
\theta_{(n+1)m} = (I - \eta M^{\top} B D_k M ) \theta_{nm} + \eta M^{\top} B D_kR + \eta M^{\top} B D_k \gamma \hat{P}
\begin{pmatrix}
\lVert \Phi_1 \theta_{nm} \rVert_{\infty} \\
\lVert \Phi_2 \theta_{nm} \rVert_{\infty} \\
\cdots \\
\lVert \Phi_{\mathcal{S}} \theta_{nm} \rVert_{\infty} \\
\end{pmatrix}.
\end{equation}

When $\theta_0 = M^{\top}y \in row\_sp(M)$ for some $y \in \mathbb{R}^k$, $\theta_t$ stays in the row space of $M$ for all $t$ and the estimated Q-values satisfy
\begin{align}
    &M \theta_{(n+1)m} = M (I - \eta M^{\top} B D_k M ) \theta_{nm} + \eta M M^{\top} B D_kR + \eta M M^{\top} B D_k \gamma \hat{P}
\begin{pmatrix}
\lVert \Phi_1 \theta_{nm} \rVert_{\infty} \\
\lVert \Phi_2 \theta_{nm} \rVert_{\infty} \\
\cdots \\
\lVert \Phi_{\mathcal{S}} \theta_{nm} \rVert_{\infty} \\
\end{pmatrix}\nonumber \\
&=  (I - \eta M M^{\top} B D_k ) M\theta_{nm} + \eta M M^{\top} B D_kR + \eta M M^{\top} B D_k \gamma \hat{P}
\begin{pmatrix}
\lVert \Phi_1 M^{\dagger}M\theta_{nm} \rVert_{\infty} \\
\lVert \Phi_2 M^{\dagger}M\theta_{nm} \rVert_{\infty} \\
\cdots \\
\lVert \Phi_{\mathcal{S}} M^{\dagger}M\theta_{nm} \rVert_{\infty} \\
\end{pmatrix}. \label{eq:Q_learning_update_rule}
\end{align}
The last line uses that $\theta_{nm} = M^{\top}y = M^{\dagger}M M^{\top} y = M^{\dagger}M \theta_{nm}$.

Compared to the TD analysis, the only change here is that instead of using next states' value estimations, we use the maximum Q-values at next states.

According to the above combined $m$ step update rule on $M \theta$ in Equation \ref{eq:Q_learning_update_rule}, we define an operator $\mathcal{T}:\mathbb{R}^{k} \to \mathbb{R}^{k}$ for $x = M\theta \in \mathbb{R}^{k}$ as
\begin{align*}
  \mathcal{T}x &= (I - \eta MM^{\top} B D_k  )  x+ \eta  M M^{\top} B D_kR + \eta M M^{\top} B D_k \gamma \hat{P}\begin{pmatrix}
\lVert \Phi_1 M^{\dagger} x \rVert_{\infty} \\
\lVert \Phi_2 M^{\dagger} x\rVert_{\infty} \\
\cdots \\
\lVert \Phi_{\mathcal{S}} M^{\dagger} x \rVert_{\infty} \\
\end{pmatrix}\\.
\end{align*}

Notice that this operator is contractive. Given any two vectors $\bar{x}$ and $x'$,
\begin{align}
& \lVert \mathcal{T} x - \mathcal{T} x' \rVert_{\infty} \nonumber \\
&=
\lVert (I-\eta M M^{\top} B D_k)(\bar{x} -  x') + \eta M M^{\top} B D_k \gamma \hat{P} [
\begin{pmatrix}
\lVert \Phi_1 M^{\dagger }\bar{x}\rVert_{\infty} \\
\lVert \Phi_2 M^{\dagger }\bar{x}\rVert_{\infty} \\
\cdots \\
\lVert \Phi_{\mathcal{S}} M^{\dagger }\bar{x} \rVert_{\infty} \\
\end{pmatrix} -
\begin{pmatrix}
\lVert \Phi_1 M^{\dagger }x'\rVert_{\infty} \\
\lVert \Phi_2 M^{\dagger }x' \rVert_{\infty} \\
\cdots \\
\lVert \Phi_{\mathcal{S}} M^{\dagger }x' \rVert_{\infty} \\
\end{pmatrix}
] \rVert_{\infty}  \nonumber \\
& \le \lVert (I - \eta MM^{\top}D_k)^m (\bar{x} -  x') \rVert_{\infty}
+ \lVert I - (I - \eta MM^{\top}D_k)^m \rVert_{\infty} \lVert \gamma \hat{P} \begin{pmatrix}
\lVert \Phi_1 M^{\dagger }(\bar{x} -  x')\rVert_{\infty} \\
\lVert \Phi_2 M^{\dagger }(\bar{x} -  x') \rVert_{\infty} \\
\cdots \\
\lVert \Phi_{\mathcal{S}} M^{\dagger }(\bar{x} -  x') \rVert_{\infty} \\
\end{pmatrix} \rVert_{\infty}\nonumber \\
& \le [ \lVert (I - \eta MM^{\top}D_k)^m\rVert_{\infty} + \gamma \lVert I - (I - \eta MM^{\top}D_k)^m \rVert_{\infty}] \lVert \Phi' M^{\dagger}\rVert_{\infty} \lVert \bar{x} - x' \rVert_{\infty}.
\end{align}
The third line uses $\eta M M^{\top} B D_k = I - (I - \eta MM^{\top}D_k)^m $ following the recursive argument in Equation \ref{eq:student_part_I}. 

Set $\bar{m} = 1+ \lceil \cfrac{\log(c-\gamma) - \log(1+\gamma)}{\log(1-\eta \lambda_k)}\rceil$ for some constant $c \in (0,1)$. As shown in Lemma \ref{thm:radius_less_one}, $\lVert (I - \eta MM^{\top}D_k)^m\rVert_{\infty} < \frac{c-\gamma}{1+\gamma}$. In this case,
the above norm can be bounded by $\frac{c-\gamma}{1+\gamma}+ \gamma [1+\frac{c-\gamma}{1+\gamma}]\lVert \bar{x} - x' \rVert_{\infty}$ and
is smaller than $c \lVert \bar{x} - x' \rVert_{\infty}$. Thus, the operator on estimated Q-values is contractive.

The dataset defines a fixed point for estimated Q-values as
\[\hat{q}^* = R + \gamma  \hat{P} \begin{pmatrix}
\lVert \Phi_1 M^{\dagger} \hat{q}^* \rVert_{\infty} \\
\lVert \Phi_2 M^{\dagger} \hat{q}^* \rVert_{\infty} \\
\cdots \\
\lVert \Phi_{\mathcal{S}} M^{\dagger} \hat{q}^*\rVert_{\infty} \\
\end{pmatrix},\]
and $\hat{q}^*$ exists uniques since the right hand side update rule is contractive.

This $\hat{q}^*(s,a)$ is also the unique fixed point of the operator $\mathcal{T}$ in the metric space $(\mathbb{R}^k,\lVert \cdot \rVert_{\infty})$ by the Banach fixed point theorem, since
\begin{align*}
  \mathcal{T} \hat{q}^* &= (I - \eta M M^{\top} B D_k  ) \hat{q}^* + \eta  M M^{\top} B D_kR + \eta M M^{\top} B D_k \gamma \hat{P}\begin{pmatrix}
\lVert \Phi_1 M^{\dagger} \hat{q}^*\rVert_{\infty} \\
\lVert \Phi_2 M^{\dagger} \hat{q}^* \rVert_{\infty} \\
\cdots \\
\lVert \Phi_{\mathcal{S}} M^{\dagger} \hat{q}^* \rVert_{\infty} \\
\end{pmatrix}\\
&=(I- \eta M M^{\top} D_k)^m \hat{q}^* + [I - (I-\eta M M^{\top} D_k)^m][R + \gamma \hat{P}\begin{pmatrix}
\lVert \Phi_1 M^{\dagger} \hat{q}^*\rVert_{\infty} \\
\lVert \Phi_2 M^{\dagger} \hat{q}^* \rVert_{\infty} \\
\cdots \\
\lVert \Phi_{\mathcal{S}} M^{\dagger} \hat{q}^* \rVert_{\infty} \\
\end{pmatrix}]\\
&= (I- \eta M M^{\top} D_k)^m \hat{q}^* + [I - (I-\eta M M^{\top} D_k)^m]\hat{q}^* \\
&= \hat{q}^* .
\end{align*}
The second equation again uses $\eta M M^{\top} B D_k = I - (I - \eta MM^{\top}D_k)^m $ following the recursive argument in Equation \ref{eq:student_part_I}. The third line uses the definition of $\hat{q}^*$.

An initial point can be expressed as $\theta_0 = M^{\dagger} M \theta_0 + (I - M^{\dagger}M) \theta_0$. 

For the part, $M^{\dagger} M \theta_0$, lying in the row space of $M$, we have $M M^{\dagger} M \theta_0$ converges uniquely to the fixed point $\hat{q}^* = M \theta^*$ where $\theta^* = M^{\top} y $ lies in the row space of $M$ for some $y \in \mathbb{R}^k$. Thus, $M^{\dagger} M \theta_t \to \theta^* = M^{\dagger} \hat{q}^*$ converges uniquely as $t \to \infty$.

The other part $(I - M^{\dagger}M) \theta_0$ is always unmodified by the over-parameterized target Q-learning update rule and thus left unchanged as $(I - M^{\dagger}M) \theta_t = (I - M^{\dagger}M) \theta_0$ for all $t$.

Therefore, $\theta_t \to M^{\dagger} \hat{q}^* + (I - M^{\dagger}M) \theta_0$ as $t \to \infty$.
\end{proof}

\subsection{Empirical Setting}\label{A:exper}

\begin{table}
\begin{center}
\begin{tabular}{ c |c |c|c }
 Algorithm & Learning Rate & Second Learning Rate & Target Update Step \\
 \hline
 TD & $0.5$ & None &None \\ 
 \hline
 Target TD & $0.997$ & None & $3$\\
 \hline
 RM & $0.8$ & None &None \\ 
 \hline
 Baird RM & $0.95$ & None &None \\ 
 \hline
 GTD2 & $0.6$ & $0.6$ &None \\
 \hline
 TDC & $0.6$ & $0.4$ &None\\
\end{tabular}
\end{center}
\caption{The table shows hyperparameters for all algorithms tuned on the Baird counterexample. All hyperparameters are found by grid search.}
\end{table}
\begin{figure}[ht]
    \centering
    \hspace*{-6mm}
    \includegraphics[width=0.5\linewidth]{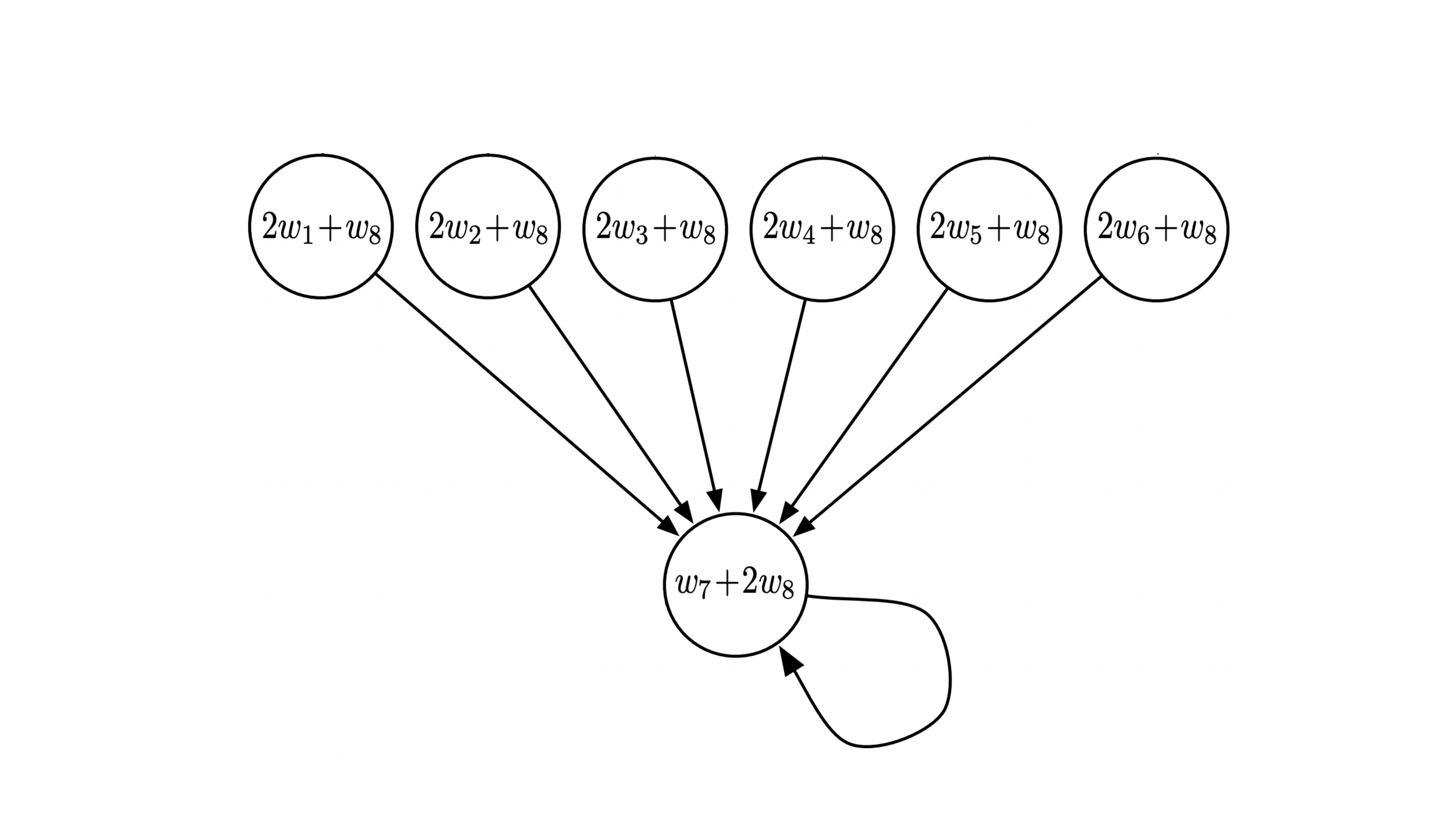}
    \vspace*{-3mm}
    \caption{The features are shown in the figure. The transition is labelled with arrows. This Baird Counterexample is a Markov Reward process and only one action is available at each state.}
    \label{fig:baird_task}
\end{figure}

\begin{figure}[ht]
    \centering
    \hspace*{-6mm}
    \includegraphics[width=0.5\linewidth]{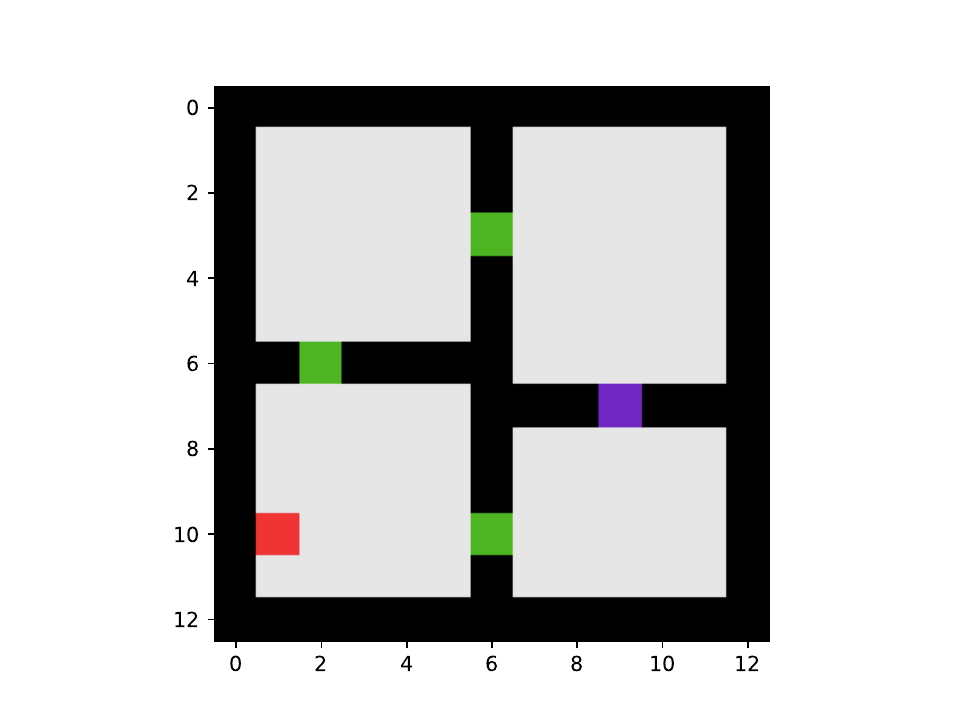}
    \vspace*{-3mm}
    \caption{Black blocks are walls which cannot be trespassed, green ones are hallways and the purple block is the terminal state with $+1$ reward. Each state has $(x,y)$ coordinate and actions include up, down, left and right.}
    \label{fig:room_task}
\end{figure}

The features for the four room concatenate one-hot encoding for $x$ and $y$ coordinate separately and the action. Then, we append the matrix $H^{\top}$, each row of length $k$, representing if state-action pairs show up in the dataset and the showing-up order, to the encoding. Thus, the dimension of features is larger than the number $k$ of state-actions in the dataset and the model is over-parameterized.

The behaviour policy is random, that is, four actions are sampled with the same probability. A human policy is given by a human player, considered as optimal by the player. The target policy is the human policy combined with $\epsilon$-exploration with $\epsilon=0.08$.

All hyperparameters are tuned with the small dataset of size $300$. Since over-parameterized TD converges on the Four Room task, then the target parameter update step is set to one and OTTD is the same as TD. All empirical results are averaged over $10$ random seeds.

\begin{table}
\begin{center}
\begin{tabular}{ c |c  }
 Algorithm & Learning Rate  \\
 \hline
 Off-Policy without Corrections & $0.95$  \\ 
 \hline
 Target Actions & $0.97$ \\
 \hline
 NIS & $0.97$  \\ 
 \hline
 IS & $0.02$ \\ 
\end{tabular}
\end{center}
\caption{The table shows hyperparameters for all algorithms tuned on the Four Room Task. All hyperparameters are found by grid search.}
\end{table}

\subsection{Comparison Between Convergence Conditions} \label{A:assump}
\begin{table*}[!htbp]
    \centering
    \begin{tabularx}{\textwidth}{Y|Y|Y}
        \hline
         Work&  Regularity Condition&Comment\\
         \hline
         Ours Thm.\ 3.2 & $\lVert NM^\dagger \rVert<1$ & It is satisfied for expected updates or a batch of complete trajectories naturally.\\
         \hline
         Lee and He Thm.\ 1 (2019)&  None& No regularization is needed for the on-policy learning. \\
         \hline
         Asadi et al.\ Prop.\ 1 (2023) & $\rho((\Phi^\top D \Phi)^{-1} (\gamma \Phi^\top D P_{\pi} \Phi)) <1$ & The condition fails on a Two-state counterexample even with expected updates.\\
         \hline
         Asadi et al.\ Prop.\ 5 (2023) & $ \frac{\lambda_{max}(\gamma \Phi^\top D P_{\pi} \Phi))}{\lambda_{min}((\Phi^\top D \Phi)} <1$ & The condition fails on a Two-state counterexample even with expected updates.\\
         \hline
         Fellows et al.\ Thm.\ 2 (2023) &  $M^\top D_k (\gamma N -M)$ has strictly negative eigenvalues & The condition is equivalent to the spectral radius less-than-one condition. Breaking this condition is the main factor behind the divergence with the deadly triad. With this assumption, the paper does not focus on the deadly triad issue.\\
         \hline
         Fellows et al.\ Thm.\ 4 (2023) &  $\lVert (\Phi^\top D \Phi)^{-1} (\gamma \Phi^\top D P_{\pi} \Phi) \rVert<1$ &  The condition fails on a Two-state counterexample even with expected updates.\\
         \hline
         Shangtong et al.\ Thm.\ 2 (2021) & Projection of the target parameter into a ball \footnotemark[2] and L2 regularization & Projection is hard to realize empirically, and L2 regularization can give a parameter predicting worse than zero values.\\
         \hline
    \end{tabularx}
    \caption{This table compares how strong the regularity conditions are to ensure convergence in the deadly triad under linear function approximation.}
    \label{tab:assump}
\end{table*}
\footnotetext[2]{The size depends on the feature norm, reward norm and the regularization weight.}

\begin{table*}[!htbp]
    \centering
    \begin{tabularx}{\textwidth}{Y|Y|Y|Y}
        \hline
         Work&  MDP &  Data Generation Distribution& Features\\
         \hline
         Ours Thm.\ 3.2 & None & None & Full rank  \\
         \hline
         Lee and He Thm.\ 1 (2019)& Ergodic under the target policy $\pi$& $s \sim d_{\pi}$ i.i.d. with $d_{\pi}(s) >0$ for all $s$ & Full rank \\
         \hline
         Asadi et al.\ Prop.\ 1 (2023) &None & None & Full rank \\
         \hline
         Asadi et al.\ Prop.\ 5 (2023) &None & None & Full rank \\
         \hline
         Fellows et al.\ (2023) Thm.\ 2  & None & $s \sim d$ i.i.d. for some off-policy distribution $d$ & $\lVert \phi(s,a)\phi(s,a)^\top \rVert$ and $\gamma \lVert \phi(s,a)\phi(s',a')^\top \rVert$ are bounded, the space of the parameter $\theta$ is convex, and variance of the update is bounded\footnotemark[3]\\
         \hline
         Fellows et al.\ (2023) Thm.\ 4  & None & $s \sim d$ i.i.d. for some off-policy distribution $d$ & $\lVert \phi(s,a)\phi(s,a)^\top \rVert$ and $\gamma \lVert \phi(s,a)\phi(s',a')^\top \rVert$ are bounded, the space of the parameter $\theta$ is convex, and variance of the update is bounded\\
         \hline
         Shangtong et al.\ Thm.\ 2 (2021) & Ergodic under the behaviour policy & Trajectory data of an infinite length& Full rank and $\lVert \Phi \rVert < C(\eta,\lVert P_{\pi}\rVert_{D_u})$\footnotemark[4]\\
         \hline
    \end{tabularx}
    \caption{Comparison of assumptions among analysis of target networks under linear function approximation.}
    \label{tab:assump_!}
\end{table*} 
\footnotetext[3]{$Var_{S\sim d, A\sim \mu,S'A' \sim P_{\pi}}(\phi(S,A)(r(s,a)+\gamma \phi(S',A')^\top \theta - \phi(S,A)^\top \theta))$ is bounded.}
\footnotetext[4]{ for some dependent constant $C$ on the regularization weight $\eta$ and transition norm.}

\begin{table*}[!htbp]
    \centering
    \begin{tabularx}{\textwidth}{Y|Y|Y}
        \hline
         Work&  Learning Rate& Target Network Hyperparameter\\
         \hline
         Ours Thm.\ 3.2 & $\eta < \frac{1}{\rho(MM^TD_k)}$ & $m \ge \bar{m}$\footnotemark[5]\\
         \hline
         Lee and He Thm.\ 1 (2019)&  Decaying learning rate $\alpha_t>0$ such that $\sum_{t=0}^{\infty} \alpha_t = \infty$ and $\sum_{t=0}^{\infty} \alpha_t^2 < \infty$ & Share the learning rate with the student or original parameter\\
         \hline
         Asadi et al.\ Prop.\ 1 (2023) & $\eta=1$ & $m=\infty$\\
         \hline
         Asadi et al.\ Prop.\ 1 (2023) & $\eta=\frac{1}{\lambda_{max}(\Phi^\top D \Phi)}$ & $m \ge 1$\\
         \hline
         Fellows et al.\ (2023) Thm.\ 2  &  Decaying learning rate $\alpha_t>0$ such that $\sum_{t=0}^{\infty} \alpha_t = \infty$ and $\sum_{t=0}^{\infty} \alpha_t^2 < \infty$ & None\\
         \hline
         Fellows et al.\ (2023) Thm.\ 4  &  $\frac{1}{\eta}>\frac{\lambda_{min}(\Phi^\top D \Phi) + \lambda_{max}(\Phi^\top D \Phi)}{2}$ & $m > \Tilde{m}$\footnotemark[6]\\
         \hline
         Shangtong et al.\ Thm.\ 2 (2021) & Decaying learning rate $\alpha_t>0$ such that $\sum_{t=0}^{\infty} \alpha_t = \infty$ and $\sum_{t=0}^{\infty} \alpha_t^2 < \infty$ & Decaying learning rate $\beta_t>0$ for the target network such that $\sum_{t=0}^{\infty} \beta_t = \infty$, $\sum_{t=0}^{\infty} \beta_t^2 < \infty$ and for some $d>0$, $\sum_{t=0}^{\infty} (\beta_t/ \alpha_t)^d < \infty$\\
         \hline
    \end{tabularx}
    \caption{Comparison of assumptions among analysis of target networks under linear function approximation.}
    \label{tab:assump_2}
\end{table*} 
\footnotetext[5]{$\bar{m} = 1+ \ceil{ \cfrac{\log(1-\gamma) - \log((1+\gamma)\sqrt{k})}{\log(1-\eta \lambda_{\mathrm{min}}(MM^TD_k))}}$ when regularizing the infinity norm of $NM^\dagger$.}
\footnotetext[6]{$\Tilde{m} = 1+  \cfrac{\log(1-\lVert \bar{J}*_{FPE}) \rVert) - \log(\lVert \bar{J}*_{FPE}) \rVert+\lVert \bar{J}*_{TD}) \rVert)}{\log(1-\eta \lambda_{\mathrm{min}}(\Phi^\top D \Phi))}$ where $\lVert \bar{J}*_{FPE}) \rVert = \lVert (\Phi^\top D \Phi)^{-1} (\gamma \Phi^\top D P_{\pi} \Phi \rVert$ and $\lVert \bar{J}*_{TD}) \rVert = \lVert I -\eta \Phi^\top D(I-\gamma P_{\pi}\Phi) \rVert$.}

\end{document}